\documentclass[times, review, 10pt]{elsarticle}

\usepackage{amssymb}
\usepackage{amsmath}
\usepackage{wrapfig} % 그림 주변에 텍스트를 흐르게 하는 패키지
\usepackage{tabularx} % 테이블을 위한 패키지, 테이블 너비 조정 가능
\usepackage{float}
\usepackage{caption}

\usepackage{color, colortbl}
\definecolor{Gray}{gray}{0.9}
\usepackage{ctable}
\usepackage{hhline}
\usepackage{multirow}

\newcolumntype{L}[1]{>{\raggedright\let\newline\\\arraybackslash\hspace{0pt}}m{#1}}
\newcolumntype{C}[1]{>{\centering\let\newline\\\arraybackslash\hspace{0pt}}m{#1}}
\newcolumntype{R}[1]{>{\raggedleft\let\newline\\\arraybackslash\hspace{0pt}}m{#1}}

\usepackage{gensymb} % \degree 를 사용할 수 있게 한다.

\usepackage{xcolor}

\biboptions{sort&compress}

\journal{Pattern Recognition}

\begin{document}

\begin{frontmatter}

%% Title, authors and addresses

%% use the tnoteref command within \title for footnotes;
%% use the tnotetext command for theassociated footnote;
%% use the fnref command within \author or \affiliation for footnotes;
%% use the fntext command for theassociated footnote;
%% use the corref command within \author for corresponding author footnotes;
%% use the cortext command for theassociated footnote;
%% use the ead command for the email address,
%% and the form \ead[url] for the home page:
%% \title{Title\tnoteref{label1}}
%% \tnotetext[label1]{}
%% \author{Name\corref{cor1}\fnref{label2}}
%% \ead{email address}
%% \ead[url]{home page}
%% \fntext[label2]{}
%% \cortext[cor1]{}
%% \affiliation{organization={},
%%             addressline={},
%%             city={},
%%             postcode={},
%%             state={},
%%             country={}}
%% \fntext[label3]{}

\title{Bidirectional Regression for Monocular 6DoF Head Pose Estimation and Reference System Alignment}

%% use optional labels to link authors explicitly to addresses:
%% \author[label1,label2]{}
%% \affiliation[label1]{organization={},
%%             addressline={},
%%             city={},
%%             postcode={},
%%             state={},
%%             country={}}
%%
%% \affiliation[label2]{organization={},
%%             addressline={},
%%             city={},
%%             postcode={},
%%             state={},
%%             country={}}

\author[kw]{Sungho Chun} %% Author name
\ead{asw9161@kw.ac.kr}

\author[dk]{Boeun Kim} %% Author name
\ead{boeun.kim@dankook.ac.kr}

\author[bi]{Hyung Jin Chang} %% Author name
\ead{h.j.chang@bham.ac.uk}

\author[kw]{Ju Yong Chang\corref{cor1}} %% Author name
\ead{jychang@kw.ac.kr}

%% Author affiliation
\affiliation[kw]{organization={Department of Electronics and Communications Engineering, Kwangwoon University}, 
                 city={Seoul},
                 country={South Korea}}
\affiliation[dk]{organization={Department of Software Engineering, Dankook University}, 
                 city={Yongin},
                 country={South Korea}}
\affiliation[bi]{organization={School of Computer Science, University of Birmingham}, 
                 city={Birmingham},
                 country={United Kingdom}}

\cortext[cor1]{Corresponding author}

%% Abstract
\begin{abstract}
%% Text of abstract
Precise six-degree-of-freedom (6DoF) head pose estimation is crucial for safety-critical applications and human–computer interaction scenarios, yet existing monocular methods still struggle with robust pose estimation. We revisit this problem by introducing TRGv2, a lightweight extension of our previous Translation, Rotation, and Geometry (TRG) network, which explicitly models the bidirectional interaction between facial geometry and head pose. TRGv2 jointly infers facial landmarks and 6DoF pose through an iterative refinement loop with landmark-to-image projection, ensuring metric consistency among face size, rotation, and depth. To further improve generalization to out-of-distribution data, TRGv2 regresses correction parameters instead of directly predicting translation, combining them with a pinhole camera model for analytic depth estimation. In addition, we identify a previously overlooked source of bias in cross-dataset evaluations due to inconsistent head center definitions across different datasets. To address this, we propose a reference system alignment strategy that quantifies and corrects translation bias, enabling fair comparisons across datasets. Extensive experiments on ARKitFace, BIWI, and the challenging DD-Pose benchmarks demonstrate that TRGv2 outperforms state-of-the-art methods in both accuracy and efficiency. Code and newly annotated landmarks for DD-Pose will be publicly available.
\end{abstract}

%%Graphical abstract
%\begin{graphicalabstract}
%\includegraphics{grabs}
%\end{graphicalabstract}

%%Research highlights
%\input{highlights.tex}

%% Keywords
\begin{keyword} 6DoF head pose estimation \sep 3D facial landmark reconstruction \sep Bidirectional interaction \sep Reference system alignment
%% keywords here, in the form: keyword \sep keyword

%% PACS codes here, in the form: \PACS code \sep code

%% MSC codes here, in the form: \MSC code \sep code
%% or \MSC[2008] code \sep code (2000 is the default)

\end{keyword}

\end{frontmatter}

%% Add \usepackage{lineno} before \begin{document} and uncomment 
%% following line to enable line numbers
%% \linenumbers

%% main text
%%

%%%%%%%%%%%%%%%%%%%%%%%%%%%%%%%%%%%%%%%%%%%%%%%%%%%%%%%%
% Introduction
%%%%%%%%%%%%%%%%%%%%%%%%%%%%%%%%%%%%%%%%%%%%%%%%%%%%%%%%
\section{Introduction}
\label{src:introduction}

A six degrees of freedom (6DoF) head pose consists of the 3D rotation and translation of the head with respect to the camera coordinate system. It provides crucial cues for understanding where a person is looking, what they are paying attention to, and which object they are interacting with. Accordingly, head pose estimation has become a core enabling technology for a wide range of human-centered applications, such as driver monitoring, human-robot collaboration, and intelligent surveillance systems. Despite the promising potential of estimating 6DoF head pose from a single image, most existing studies~\cite{caoECCV2022towards, hempel2024toward, cobo_pr24} have primarily focused on head rotation, while head translation has received relatively less attention. However, estimating head translation from a monocular image is a highly challenging task. The main difficulty arises from the strong entanglement between facial size information and head translation---without knowing one, it is difficult to accurately estimate the other. As a result, methods~\cite{zielonka2022mica, guo2022jmlr, 2023_tip_perspnet} that attempt to estimate head translation by first reconstructing face geometry often suffer from translation ambiguity, while those~\cite{albiero2021img2pose, roth2023intrapose} that attempt to predict head translation without facial size information are prone to geometry ambiguity.

Existing 6DoF head pose estimation methods~\cite{albiero2021img2pose, zielonka2022mica, 2023_tip_perspnet, guo2022jmlr, roth2023intrapose} can be broadly categorized into optimization-based approaches and direct regression approaches. However, these methods do not effectively address the aforementioned challenge. Optimization-based approaches~\cite{zielonka2022mica, 2023_tip_perspnet, guo2022jmlr, xu2024multi} typically reconstruct face geometry from the input image first, and then estimate the head pose through geometric alignment. However, this strategy operates in a unidirectional manner, from face geometry to head pose, and does not incorporate any feedback from the head pose back to the geometry. In other words, these methods do not explicitly model depth information during the geometry reconstruction process, which makes it difficult to accurately estimate the actual facial size. Consequently, the optimization process based on such inaccurate geometry significantly degrades the accuracy of the subsequent head translation estimation. In contrast, direct regression methods such as img2pose~\cite{albiero2021img2pose, roth2023intrapose} aim to directly predict head rotation and translation from an RGB image without modeling facial geometry. However, directly regressing the 6DoF head pose from an image is a highly non-linear problem, and ignoring facial geometry---which could serve as a depth-related constraint---can lead to degraded translation estimation performance.

\begin{figure*}[t]
\centering
\includegraphics[width=1.0\linewidth]{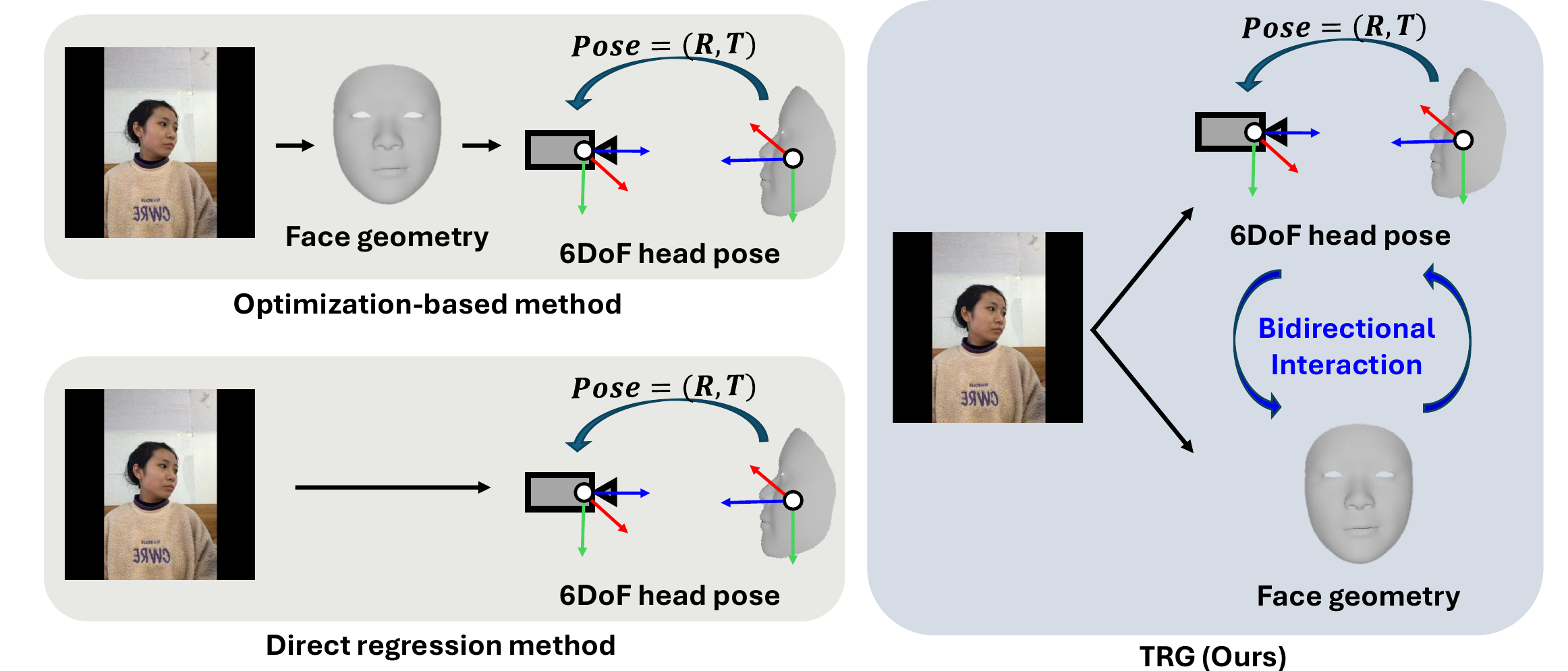}
\caption{Illustration of different methods for inferring 6DoF head pose. Optimization-based methods~\cite{zielonka2022mica, guo2022jmlr, 2023_tip_perspnet} first predict the face geometry and then calculate the head pose sequentially. The regression-based approach~\cite{albiero2021img2pose} directly regresses the head pose from the input image. In contrast, the proposed method jointly estimates both the face geometry and the head pose to fully exploit the synergy between them.}
\label{fig:main_idea}
\end{figure*}

To overcome these limitations, we propose the \emph{{\textbf{T}}ranslation, {\textbf{R}}otation, and {\textbf{G}}eometry network (TRG)}---a regression-based model for 6DoF head pose estimation that \emph{explicitly designs a bidirectional interaction between facial geometry and head pose}. As illustrated in Fig.~\ref{fig:main_idea}, unlike previous pipelines~\cite{zielonka2022mica, guo2022jmlr, 2023_tip_perspnet} that rely solely on geometry-to-pose inference, TRG incorporates feedback from pose back to geometry, enabling the mutual refinement of both components. The proposed method first generates initial predictions for head pose and geometry, and then iteratively refines them so that the two elements can complement and enhance each other. Within each iteration, TRG performs a landmark-to-image projection to exploit the complementary relationship between facial size and 6DoF head pose. Specifically, it projects the predicted 3D facial geometry onto the image plane using the current head pose estimate, and then samples image features at the projected locations. The more accurate the head pose and geometry are, the more precisely the landmarks can be projected onto the correct 2D positions, and the features sampled at those locations enable more accurate estimation of both head pose and geometry. Through experiments, we demonstrate that this bidirectional interaction mechanism, grounded in landmark-to-image projection, leads to improved accuracy in both head translation and rotation.

% fig 1
% fig 1

Another challenge in estimating head translation from a single image is the difficulty of generalizing regression-based methods to out-of-distribution data. This is because the head translations---especially the depth values---contained in training data typically span a limited range, whereas those in real-world test data can vary widely, from a few centimeters to several meters. In contrast, simple geometry-based methods grounded in the pinhole camera model are less affected by this generalization issue. Assuming that the camera's focal length is known, one can derive a closed-form formula for head depth estimation using the subject's real-world head size (e.g., 20 cm) and the bounding box that tightly encloses the head in the image. However, real-world variation in head sizes and inaccuracies in bounding box detection limit the reliability of this approach. To address these issues, we propose a method that \emph{retains the geometric formulation based on the pinhole camera model, while explicitly estimating the variation in head size and bounding box position}. Rather than directly regressing the unbounded translation vector, we regress correction parameters for the variables used in the geometric model. This approach is not only more tractable but also empirically shown to yield better performance.

In the context of head translation estimation, we found that the evaluation protocol itself requires reconsideration. To properly assess a model's generalization ability, datasets with different distributions are often used for training and testing. However, inconsistencies in head pose annotation schemes across datasets can distort the accuracy and fairness of performance evaluations for various methods.
For example, head pose is commonly annotated using a dataset-specific 3D mesh, but the definition of the head center is not consistently standardized across datasets. This inconsistency in head center definition introduces bias in translation evaluation, making it difficult to construct accurate and fair benchmarks. To address this issue, we propose a \emph{reference system alignment strategy} that quantifies and corrects the translation bias between datasets, thereby enabling more valid and consistent performance evaluation. Specifically, we compute the head center offset between different mesh structures and apply this offset as a correction term during model evaluation. This approach improves the accuracy and consistency of cross-dataset comparisons and helps identify models that may have been undervalued due to annotation mismatches despite their strong actual performance. To the best of our knowledge, this is the first work to systematically define and resolve the domain misalignment problem in head translation evaluation.

The main contributions of this work are summarized as follows:
\begin{itemize}
\item We propose TRG, a novel framework for regressing 6DoF head pose from a single image, which explicitly models the bidirectional interaction between head pose and facial geometry.
\item To enforce consistency between the predicted pose and facial landmarks, we introduce a landmark-to-image projection strategy that improves both translation and rotation accuracy.
\item Our correction parameter regression strategy for translation estimation enables accurate predictions and demonstrates strong generalization to out-of-distribution data.
\item The head pose-aware landmark estimation architecture in TRG achieves accurate 3D reconstruction performance even under strong perspective distortion (e.g., selfie views).
\item To the best of our knowledge, this is the first study to explicitly identify and correct bias caused by domain gaps in cross-dataset head translation evaluation. We propose a reference system alignment strategy to mitigate this bias.
\item TRG achieves state-of-the-art results on three public benchmarks---ARKitFace~\cite{2023_tip_perspnet}, BIWI~\cite{fanelli2013biwi}, and DD-Pose~\cite{roth2019iv}---demonstrating superior accuracy and generalization ability.
\end{itemize}

An earlier version of this work was presented at ECCV 2024~\cite{chun2024trg}. In this journal version, we significantly extend the original study in the following three major aspects: First, we identify and correct a previously overlooked evaluation bias caused by discrepancies in head mesh topologies across datasets. By applying the proposed reference system alignment strategy to both TRG and existing methods, we enable more reliable and fair cross-dataset comparisons. Second, we demonstrate the robustness of TRG on the DD-Pose benchmark, which includes challenging scenarios such as occlusion, illumination variation, and truncation. To facilitate future research, we also release newly annotated sparse 2D facial landmarks for this dataset. Third, to improve the practical applicability of our method, we propose a simplified variant, TRGv2, with significantly reduced computational cost. While the original conference version utilizes three refinement iterations and multi-scale feature aggregation, the journal version adopts a more lightweight design with only two iterations and single-scale feature extraction. Despite this simplification, TRGv2 achieves performance comparable to the original model, while offering faster inference and lower GPU memory usage.

%%%%%%%%%%%%%%%%%%%%%%%%%%%%%%%%%%%%%%%%%%%%%%%%%%%%%%%%
% Related works
%%%%%%%%%%%%%%%%%%%%%%%%%%%%%%%%%%%%%%%%%%%%%%%%%%%%%%%%
\section{Related work}
\label{src:related_work}

\subsection{3D head rotation estimation}

A substantial body of work has focused exclusively on head rotation estimation, contributing significantly to advances in computer vision. Prior studies have explored various rotation representations~\cite{hempel20226d, hsu2018quatnet} and have proposed methods to reduce the uncertainty associated with head rotation~\cite{caoECCV2022towards}. Some works~\cite{hempel20226d, cobo_pr24} have further developed to design loss functions specifically optimized for head rotation estimation. In addition, there have been attempts to unify face detection and head rotation estimation into a single framework~\cite{zhou2023directmhp}. In another line of research, some existing studies have either relied on face geometry to predict head rotation or obtained head rotation as a byproduct of reconstructing facial geometry~\cite{ranjan2019hyperface, wu2021synergy, feng2021deca}. Unlike the methods mentioned above, recent efforts have also focused on head rotation estimation in challenging scenarios, such as full-range head rotation~\cite{zhou2024semi, hempel2024toward, cobo_pr24} and head rotation estimation under occlusion~\cite{celestino20232d}. However, these studies differ significantly from ours in that they estimate only head rotation and do not address 3D head translation.

\subsection{6DoF head pose estimation}

6DoF head pose estimation studies can largely be categorized into regression-based approaches and optimization-based approaches. Existing regression-based methods~\cite{albiero2021img2pose, roth2023intrapose} compute head translation from region proposals and adopt a local-to-global transformation strategy to convert the estimated local pose to the global image space. However, these methods do not incorporate any facial geometry information, which can further aggravate the issue of depth ambiguity. TRGv2 also follows a regression-based approach; however, unlike previous works such as \cite{albiero2021img2pose, roth2023intrapose}, it jointly infers face geometry along with the 6DoF head pose, explicitly leveraging the synergy between the two components, which marks a clear distinction from prior approaches.

Optimization-based methods for 6DoF head pose estimation~\cite{zielonka2022mica, guo2022jmlr, 2023_tip_perspnet, xu2024multi} adopt a sequential pipeline, in which face geometry is predicted first, followed by head pose inference via geometric alignment. While this approach is intuitive, it suffers from inherent limitations in reconstructing metric-scale geometry, primarily due to depth ambiguity. As a result, when such inaccurate geometric priors are used to estimate 6DoF head pose, the overall translation accuracy often degrades significantly. In contrast, TRGv2 introduces a novel formulation that incorporates a bidirectional interaction mechanism between 6DoF head pose and 3D facial geometry. Unlike previous optimization-based methods, TRGv2 explicitly integrates head pose cues during face geometry estimation, enabling more reliable and geometry-aware predictions.

\subsection{Landmark-to-image projection}

Our landmark-to-image projection strategy is inspired by PyMAF~\cite{zhang2021pymaf}, but PyMAF and TRGv2 differ in three key aspects. First, PyMAF aims to reconstruct a human mesh from a monocular image. In other words, it focuses on predicting human identity parameters and joint rotations. Unlike TRGv2, however, it does not attempt to recover the subject at a real-world scale or to estimate the distance between the human and the camera. Second, the motivation behind using landmark-to-image projection also differs. PyMAF adopts this strategy based on the assumption that if the predicted joint rotations are accurate, the human mesh will align well with the image. In contrast, TRGv2 uses landmark-to-image projection to leverage the strong entanglement between face geometry and 6DoF head pose.
Third, from a structural perspective, PyMAF is based on weak-perspective projection, whereas TRGv2 uses perspective projection to compute translation. The use of perspective projection requires careful consideration for generalization to out-of-distribution scenarios, which leads TRGv2 to adopt a translation estimation method grounded in geometric modeling. This marks a clear architectural distinction between TRGv2 and PyMAF.

\subsection{Reference system alignment}

There exists a critical issue of reference system misalignment between different datasets. In particular, when the reference systems of the training and test datasets differ, it becomes challenging to accurately assess the performance of models. A prior study by Cobo~\cite{cobo_pr24} pointed out this problem by analyzing the difference in head rotation reference systems between the training and test datasets. However, their work focused exclusively on head rotation and did not consider issues related to head translation. Cobo's method attributes the prediction bias observed across entire frames to differences in reference systems between datasets. While this perspective is insightful, such an approach doesn't disentangle model bias errors from reference system discrepancies. This can lead to inconsistent calibration values for the models being evaluated.

In contrast, our method, to the best of our knowledge, is the first to highlight that head translation is also affected by reference system misalignment. Moreover, unlike Cobo's method, our approach doesn't rely on model predictions for alignment. This ensures consistent calibration values regardless of the model being evaluated.

%%%%%%%%%%%%%%%%%%%%%%%%%%%%%%%%%%%%%%%%%%%%%%%%%%%%%%%%
% Proposed methods
%%%%%%%%%%%%%%%%%%%%%%%%%%%%%%%%%%%%%%%%%%%%%%%%%%%%%%%%
\section{Proposed method}
\label{src:method}

% fig 2
\begin{figure*}[t]
\centering
\includegraphics[width=1.0\linewidth]{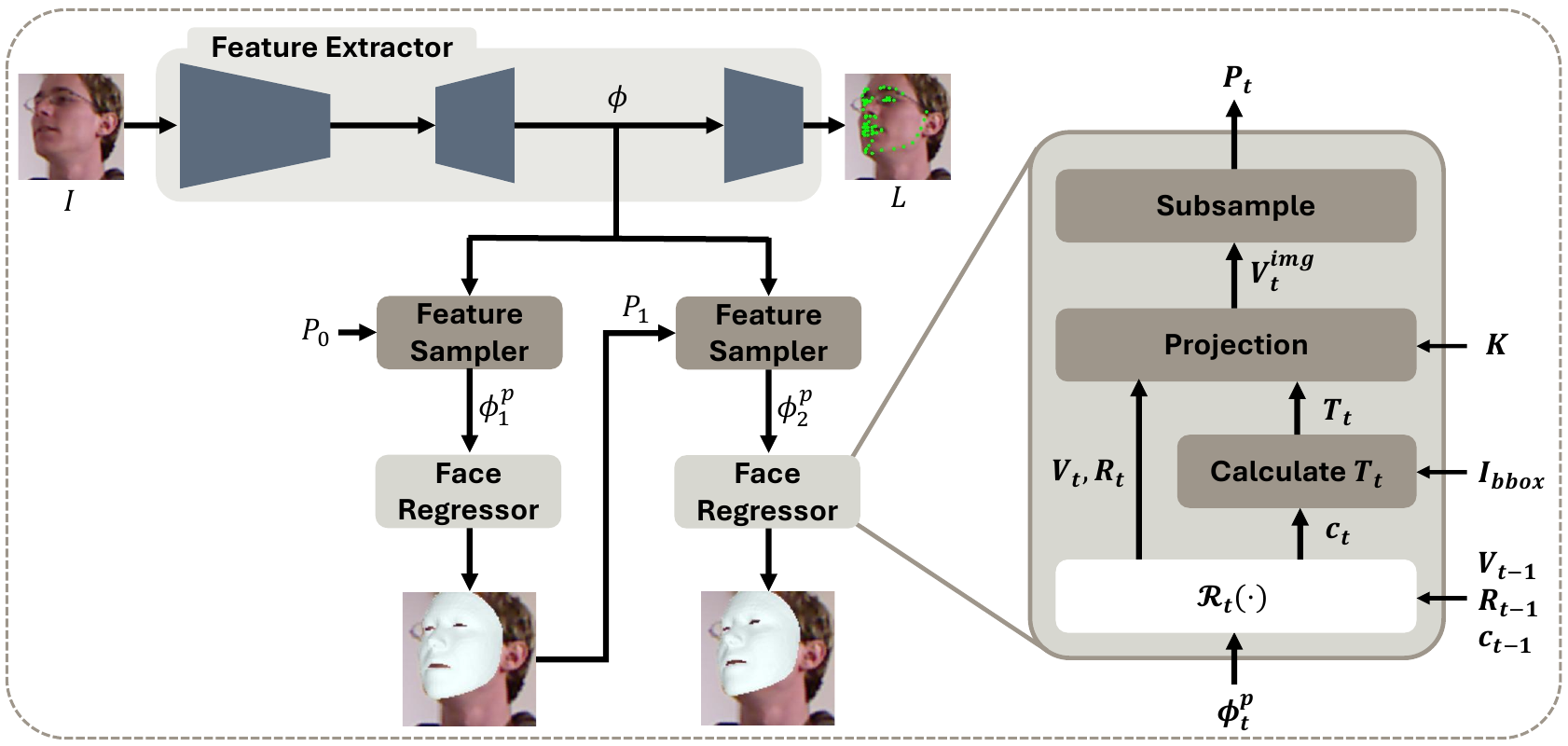}
\caption{Overall pipeline of TRGv2. TRGv2 is designed with a focus on the bidirectional interaction between face geometry and 6DoF head pose. It first performs initial predictions of face geometry and head pose, and then refines these estimates using a structure that enforces metric consistency between the two through a landmark-to-image projection strategy.}
\label{fig:overview}
\end{figure*}

%-------------------------------------------------------------------------
\subsection{Model architecture}

\subsubsection{Overview of the proposed model}

From a single image $I\in\mathbb{R}^{3\times{}H\times{}W}$, TRGv2 iteratively regresses head translation $\{T_t\in\mathbb{R}^{3}\}_{t=1}^2$ and rotation $\{R_t\in\mathbb{R}^{6}\}_{t=1}^2$, while simultaneously producing additional dense 3D landmarks $\{V_t\in\mathbb{R}^{3\times{}N^{V}}\}_{t=1}^{2}$. Fig.~\ref{fig:overview} illustrates the comprehensive structure of TRGv2, which comprises a feature extractor that generates feature maps $\phi\in\mathbb{R}^{256\times{}\frac{H}{8}\times{}\frac{W}{8}}$ from $I$; a feature sampler that extracts a landmark-aligned feature vector $\phi_t^p\in\mathbb{R}^{5N_{t-1}^P}$ from the feature map $\phi$; and a face regressor that predicts head translation $T_t$, rotation $R_t$, and dense landmarks $V_t$ from $\phi_t^p$. Here, $N_{t-1}^P$ and $N^{V}$ denote the number of sampling points $P_{t-1}\in\mathbb{R}^{2\times{}N_{t-1}^P}$ and the number of 3D dense landmarks $V_t$, respectively. Each of these components---feature extractor, feature sampler, and face regressor---is described in detail in Sections~\ref{sec:feature_extractor},~\ref{sec:feature_sampling}, and~\ref{sec:face_regressor}, respectively. Section~\ref{sec:calc_head_transl} further elaborates on how head translation is computed based on a simple geometric formulation derived from the pinhole camera model.

%-------------------------------------------------------------------------
\subsubsection{Feature extractor}
\label{sec:feature_extractor}

The feature extractor computes a feature map $\phi$ and 2D sparse landmarks $L\in\mathbb{R}^{2\times{}N^L}$ from a single image $I$. Here, $N^L$ denotes the number of sparse landmarks. The feature extractor comprises ResNet18~\cite{2016_He}, two deconvolution layers, a convolution block, and a soft-argmax operation~\cite{2018_Sun}. ResNet18 is initialized with pre-trained weights on ImageNet~\cite{imagenet2009cvpr} and used with its final classification and pooling layers removed. The feature map $\phi$ is computed by the deconvolution layers and fed into the feature samplers. Additionally, $\phi$ is further transformed into 2D heatmaps through the convolution block. The convolution block consists of a 3×3 convolutional layer, followed by batch normalization, a Leaky ReLU activation~\cite{maas2013leakyrelu}, and a 1×1 convolutional layer. The soft-argmax operation then predicts $L$ from the resultant heatmaps. These predicted landmarks, along with the ground-truth landmarks $L^{*}\in\mathbb{R}^{2\times{}N^L}$, are incorporated into the loss function.

%-------------------------------------------------------------------------
\subsubsection{Feature sampler}
\label{sec:feature_sampling}

The feature sampler computes the landmark-aligned feature vector $\phi_t^p\in\mathbb{R}^{5N_{t-1}^P}$ from the feature map $\phi$ and the corresponding sampling points $P_{t-1}\in\mathbb{R}^{2\times{}N_{t-1}^P}$. The sampling points $P_{t-1}$ are used to extract point-wise features from the feature map $\phi$. Here, $P_0$ is set to 2D grid coordinates. For $t>0$, $P_t$ is computed using the $t$-th face regressor. The procedure for deriving these sampling points is described in Section~\ref{sec:face_regressor}.

The point-wise feature vector $\phi(p_{t-1,n})\in\mathbb{R}^{256}$ is obtained using bilinear sampling at the location specified by the point $p_{t-1,n}\in\mathbb{R}^{2}$ on $\phi$. Here, $p_{t-1,n}$ denotes the $n$-th column vector of the sampling points $P_{t-1}$. The $N_{t-1}^P$ point-wise features, denoted as $\{\phi(p_{t-1,n})\}_{n=1}^{N_{t-1}^P}$, are then transformed into 5D vectors using a dimension reduction layer $\mathcal{F}(\cdot)$. These vectors are subsequently concatenated to form the landmark-aligned feature vector $\phi_t^p$:
\begin{equation}
\label{eq:mesh_aligned_feature}
    \phi_t^p=\text{concat}(\{\mathcal{F}(\phi(p_{t-1,n}))\}_{n=1}^{N_{t-1}^P}),
\end{equation}
where $\text{concat}(\cdot)$ denotes concatenation. The dimension reduction layer, $\mathcal{F}(\cdot)$, is implemented as a multilayer perceptron (MLP), which comprises three fully connected layers and two Leaky ReLU activations~\cite{maas2013leakyrelu, zhang2021pymaf}. The resulting landmark-aligned feature vector $\phi_t^p$ is then fed into the face regressor.

%-------------------------------------------------------------------------
\subsubsection{Face regressor}
\label{sec:face_regressor}

The face regressor comprises (i) an MLP, $\mathcal{R}_t(\cdot)$, that estimates the head rotation, bounding box correction parameters, and dense landmarks $\Theta_t=\{R_t\in\mathbb{R}^{6}, c_t\in\mathbb{R}^{3}, V_t\in\mathbb{R}^{3\times{}N^V}\}$; (ii) a module that computes the head translation $T_t=\{T_t^x,T_t^y,T_t^z\}\in\mathbb{R}^{3}$ based on the bounding box information $I_{\text{bbox}}=\{\frac{\tau^{x,\text{bbox}}}{f}, \frac{\tau^{y,\text{bbox}}}{f},\frac{b}{f}\}\in\mathbb{R}^{3}$ and the correction parameter $c_t=\{s_t,\tilde{\tau}_t^{x,\text{face}},\tilde{\tau}_t^{y,\text{face}}\}$; and (iii) a perspective projection function that calculates the image coordinates of the dense landmarks $V_t^{img}\in\mathbb{R}^{2\times{}N^V}$ and the sampling points $P_t$. Here, $V_t$ and $R_t$ denote the 3D coordinates of the dense landmarks defined in the head space and the head rotation expressed in a 6D representation~\cite{Zhou_2019_CVPR}, respectively. $T_t^x$, $T_t^y$, and $T_t^z$ represent the head translations along the $x$-, $y$-, and $z$-axes in the camera space, respectively. $\tau^{x,\text{bbox}}$, $\tau^{y,\text{bbox}}$, $b$, and $f$ denote the $x$- and $y$-coordinates of the bounding box center relative to the center of the uncropped image, the size of the bounding box, and the focal length, respectively. $s_t$, $\tilde{\tau}_t^{x,\text{face}}$, and $\tilde{\tau}_t^{y,\text{face}}$ respectively denote the bounding box scale factor and the normalized offsets of the head center relative to the bounding box center along the $x$- and $y$-axes.

The MLP $\mathcal{R}_t(\cdot)$ estimates the residual for calculating $\Theta_t$ from the landmark-aligned feature $\phi_t^p$, the previously iterated output $\Theta_{t-1}^{sub}=\{R_{t-1}, c_{t-1}, V_{t-1}^{sub}\in\mathbb{R}^{3\times{}305}\}$, and the bounding box information $I_{\text{bbox}}$~\cite{zhang2021pymaf, li2022cliff}. $\Theta_t$ is computed by adding the residual estimated by $\mathcal{R}_t(\cdot)$ to $\Theta_{t-1}$. $V_{t-1}^{sub}$ represents the landmarks obtained by subsampling $V_{t-1}$~\cite{ranjan2018coma}. The use of $V_{t-1}^{sub}$ for $\mathcal{R}_t(\cdot)$ instead of $V_{t-1}$ reduces the redundancy of the dense landmarks, which improves the performance of the proposed model~\cite{chun2023vha, lin2021metro, cho2022fastmetro}. The predicted $c_t$ is then converted to $T_t$ using the proposed geometric model-based method, as detailed in Section~\ref{sec:calc_head_transl}.

Finally, the image coordinates of the dense landmarks, $V_t^{img}$, are obtained by projecting $V_t$, as follows:
\begin{equation}
\label{eq:perspective_projection}
    V_t^{img}=\Pi(V_t, R_t,T_t,K),
\end{equation}
where $\Pi(\cdot)$ and $K \in \mathbb{R}^{3 \times 3}$ denote the perspective projection and the intrinsic camera matrix, respectively. The sampling points $P_t$ are obtained by subsampling $V_t^{img}$.

%-------------------------------------------------------------------------
\subsubsection{Geometric model-based translation estimation}
\label{sec:calc_head_transl}

To compute the depth $T_t^z$ using a pinhole camera model, both the actual physical size of a human face and its corresponding size in the image are required. We approximate the physical face size with a canonical square template $B$ whose side length is initially set to 0.2 meters. The projection of this square onto the image plane defines a bounding box, and its side length in pixels is denoted as $b$. Because the canonical size 0.2 meters may differ from an individual’s actual face size, the MLP $\mathcal{R}_t(\cdot)$ additionally predicts a scale factor $s_t$ to adjust $B$ before inferring depth. Furthermore, $\mathcal{R}_t(\cdot)$ is responsible for determining normalized offsets $\tilde{\tau}_t^{x,\text{face}}, \tilde{\tau}_t^{y,\text{face}}$ of the head center. These offsets represent the values obtained by normalizing the image-space translation from the bounding box center to the head center with $b$. The calculation of $T_t$ from $c_t$ and $I_{\text{bbox}}$ can be written as:
\begin{equation}
\label{eq:calc_translation}
    \begin{aligned}
        &T_t^x=\frac{0.2s_t}{b}\tau^{x,\text{bbox}}+0.2s_{t}\tilde{\tau}_t^{x,\text{face}}, \\
        &T_t^y=\frac{0.2s_t}{b}\tau^{y,\text{bbox}}+0.2s_{t}\tilde{\tau}_t^{y,\text{face}}, \\
        &T_t^z=\frac{0.2s_t}{b}f.
    \end{aligned}
\end{equation}

% fig 3
%%%%%%%%%%%%%%%%%%%%%%%%
% fig 3
% Derivation of how to calculate translation from bounding box correction parameters
\begin{figure}[t]
\centering
\includegraphics[width=1.0\linewidth]{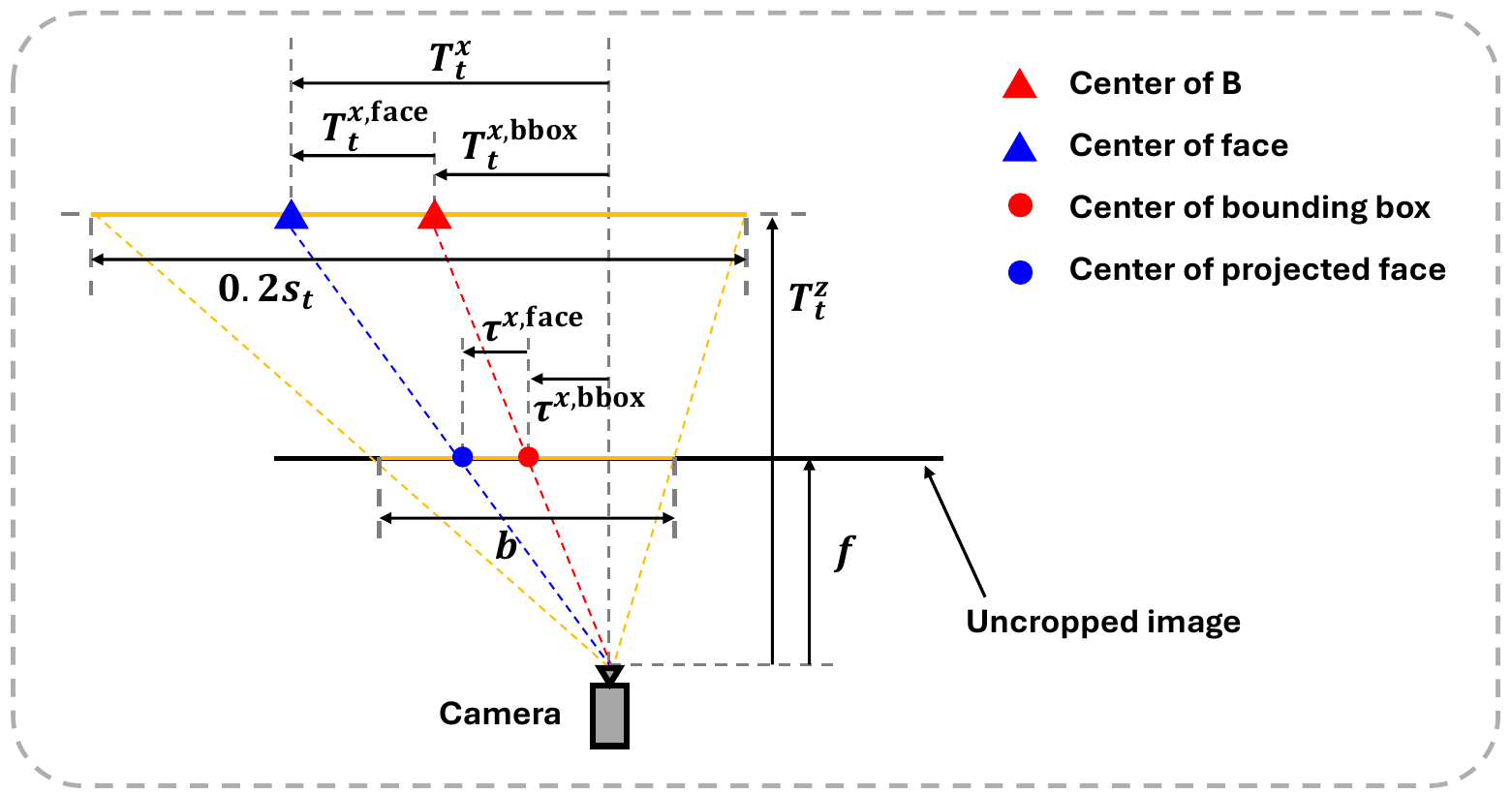}
\caption{Illustration of the calculation of head translation $T_t$ from the correction parameters $c_t$ and the bounding box information $I_\text{bbox}$. Best viewed in color.}
\label{fig:calc_translation}
\end{figure}
%%%%%%%%%%%%%%%%%%%%%%%%

Fig.~\ref{fig:calc_translation} illustrates how the head translation $T_t$ is derived from the correction parameters $c_t$ and the bounding box information $I_{\text{bbox}}$, leading to the following intermediate relationships:
\begin{equation}
\label{eq:calc_trans_1}
    \frac{T_t^z}{f}=\frac{0.2s_t}{b} \iff T_t^z=\frac{0.2s_t}{b}f,
\end{equation}
\begin{equation}
\label{eq:calc_trans_2}
    \frac{\tau_t^{x,\text{face}}}{b}=\frac{T_t^{x,\text{face}}}{0.2s_t} \iff T_t^{x,\text{face}}=\frac{0.2s_t}{b}\tau_t^{x,\text{face}},
\end{equation}
\begin{equation}
\label{eq:calc_trans_3}
    \frac{\tau^{x,\text{bbox}}}{f}=\frac{T^{x,\text{bbox}}}{T_t^z},
\end{equation}
\begin{equation}
\label{eq:calc_trans_4}
    T_t^x=T^{x,\text{bbox}}+T_{t}^{x,\text{face}}.
\end{equation}
Substituting Eq.~(\ref{eq:calc_trans_1}) into Eq.~(\ref{eq:calc_trans_3}) yields:
\begin{equation}
\label{eq:calc_trans_5}
    T^{x,\text{bbox}}=\frac{0.2s_t}{b}\tau^{x,\text{bbox}}.
\end{equation}
Inserting Eqs.~(\ref{eq:calc_trans_2}) and (\ref{eq:calc_trans_5}) into Eq.~(\ref{eq:calc_trans_4}) gives:
\begin{equation}
\label{eq:calc_trans_6}
    T_t^{x}=\frac{0.2s_t}{b}\tau^{x,\text{bbox}}+\frac{0.2s_t}{b}\tau_t^{x,\text{face}},
\end{equation}
where $\tau_t^{x,\text{face}}$ represents the $x$-axis image coordinate of the head center relative to the bounding box center. Its normalized value, $\tilde{\tau}_t^{x,\text{face}}$, is obtained by dividing $\tau_t^{x,\text{face}}$ by $b$, indicating that $\tau_t^{x,\text{face}} = b\tilde{\tau}_t^{x,\text{face}}$. Substituting $b\tilde{\tau}_t^{x,\text{face}}$ for $\tau_t^{x,\text{face}}$ in Eq.~(\ref{eq:calc_trans_6}) leads to Eq.~(\ref{eq:calc_translation}). The calculation of $T_t^y$ follows the same procedure used for $T_t^x$.

%-------------------------------------------------------------------------
\subsection{Loss functions}
\label{sec:loss}

We detail the loss functions employed to train TRGv2, ensuring accurate estimation of face geometry and head pose. The training process uses multiple losses to supervise dense landmarks and head rotation. For dense landmarks, we define the head-space coordinate loss $\mathcal{L}_{\text{head}}$, camera-space coordinate loss $\mathcal{L}_{\text{cam}}$, and image-space coordinate loss $\mathcal{L}_{\text{img}}$. To ensure precise rotation prediction, a head rotation loss $\mathcal{L}_{\text{rot}}$ is also adopted. At each iteration, these losses are progressively weighted by a factor $2^{t-2}$:
\begin{equation}
\label{eq:loss_1}
    \begin{aligned}
        &\mathcal{L}_{\text{head}}=\sum_{t=1}^{2}2^{t-2}(\frac{1}{N^V}\sum_{n=1}^{N^V}\|V_{t,n}-V_{n}^{*}\|_{1}),\\
        &\mathcal{L}_{\text{cam}}=\sum_{t=1}^{2}2^{t-2}(\frac{1}{N^V}\sum_{n=1}^{N^V}\|V_{t,n}^{cam}-V_{n}^{*,cam}\|_{1}), \\
        &\mathcal{L}_{\text{img}}=\sum_{t=1}^{2}2^{t-2}(\frac{1}{N^V}\sum_{n=1}^{N^V}\|V_{t,n}^{img}-V_{n}^{*,img}\|_{1}), \\
        &\mathcal{L}_{\text{rot}}=\sum_{t=1}^{2}2^{t-2}(\|R_t^{mat}-R^{*,mat}\|_{F}),
    \end{aligned}
\end{equation}
where $*$ and $V_{t,n}$ represent the ground truth and the $n$-th column vector of $V_t$, respectively. $V_t^{cam}=R_{t}^{mat}V_{t}+T_t\in\mathbb{R}^{3\times{}N^V}$ represents the camera-space coordinates of the $t$-th dense landmarks. $R_t^{mat}\in\mathbb{R}^{3\times{}3}$ represents the 3D head rotation in matrix form, and $\|\cdot\|_F$ denotes the Frobenius norm.

Additionally, to improve the quality of the feature map, we apply the sparse 2D landmark loss $\mathcal{L}_{L}$ to the predicted landmarks $L$ as follows:
\begin{equation}
\label{eq:loss_edge}
    \mathcal{L}_{L}=\frac{1}{N^L}\sum_{n=1}^{N^L}\|L_n-L_n^*\|_1.
\end{equation}

Finally, the overall training objective can be written as:
\begin{equation}
\label{eq:loss_lmk} \mathcal{L}=\delta_{3D}(\lambda_{\text{head}}\mathcal{L}_{\text{head}}+\lambda_{\text{cam}}\mathcal{L}_{\text{cam}})+(1-\delta_{3D})\lambda_{\text{img}}\mathcal{L}_{\text{img}}+\lambda_{\text{rot}}\mathcal{L}_{\text{rot}}+\lambda_{L}\mathcal{L}_{L},
\end{equation}
where the weights $\lambda$ control the contribution of each term, and $\delta_{3D} \in \{0,1\}$ indicates the availability of 3D ground truth.

% fig 4
%%%%%%%%%%%%%%%%%%%%%%%%
% fig 4
% Face topology, Reference system alignment
\begin{figure}[t]
\centering
\includegraphics[width=1.0\linewidth]{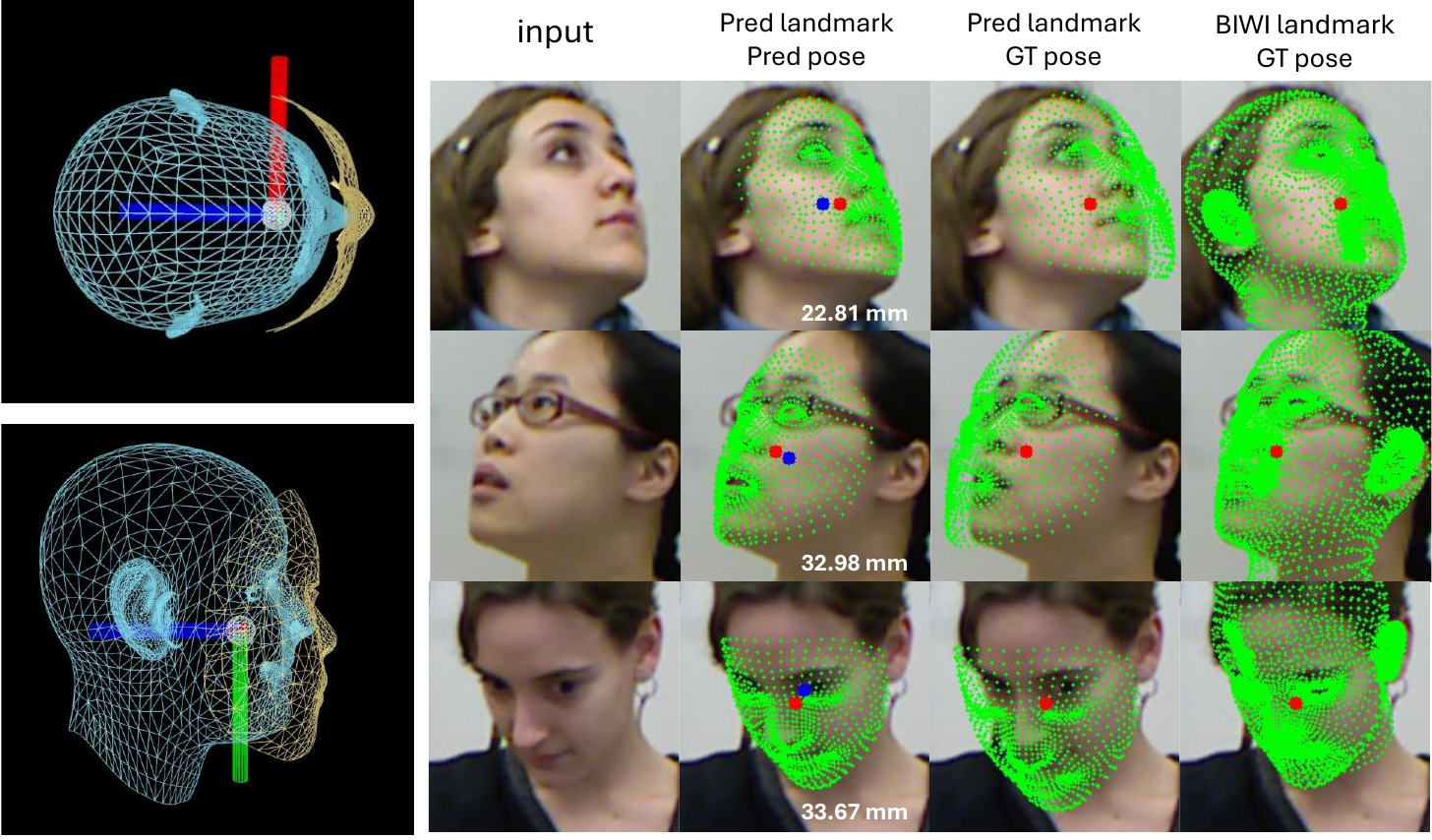}
\caption{\textbf{Left}: Visualization of the BIWI face mesh (blue) and the ARKitFace face mesh (gold), aligned at the same head center. \textbf{Right}: Predicted 3D dense landmarks and head pose from a model trained on the ARKitFace training data. The BIWI landmarks denote the 3D dense landmarks provided by the BIWI dataset. The white numbers in the second column indicate the translation error ($\text{MAE}_t$). The red and blue dots represent the ground-truth and predicted head centers, respectively.}
\label{fig:face_topology}
\end{figure}
%%%%%%%%%%%%%%%%%%%%%%%%

%-------------------------------------------------------------------------
\subsection{Reference system alignment}
\label{sec:reference_align}

Head translation refers to the coordinates of the head center defined in the camera coordinate system. Therefore, if different datasets adopt inconsistent definitions of the head center, translation bias can arise during cross-dataset evaluation. In practice, 6DoF head pose datasets~\cite{2023_tip_perspnet, fanelli2013biwi} often use different head center definitions because each dataset constructs its head pose annotations based on distinct face mesh topologies. The left panel of Fig.~\ref{fig:face_topology} illustrates that the position of the head center varies across datasets. Since the head center is defined differently for each face topology, even when the meshes are visualized after aligning their head centers, the two meshes do not perfectly overlap.

To further illustrate how translation bias can lead to misleading evaluations, the right panel of Fig.~\ref{fig:face_topology} shows an example. In the second column, the projected landmarks appear well aligned with the face in the image. However, despite this visual alignment, there remains a significant translation error when evaluated quantitatively. In contrast, in the third column, the landmarks appear misaligned in the image, but since the ground-truth translation is used, there is no translation error. The fourth column shows BIWI landmarks projected using the ground-truth head pose. Although both the third and fourth columns use the same ground-truth head pose for projection, differences in head center definitions cause the ARKitFace landmarks to appear misaligned while the BIWI landmarks align well with the image.

To remove translation bias $\Delta{T} \in \mathbb{R}^3$ during cross-dataset evaluation, we quantitatively measure the discrepancy between head centers. This is achieved by aligning the two face mesh topologies and computing the offset between their head centers. To perform the alignment, we use a subset of facial landmarks corresponding to the eyes, nose, and mouth, and adopt a gradient-based fitting method~\cite{pavlakos2019smplify-x, zhang2021mojo}. The translation bias $\Delta{T}$ is then computed by solving the following optimization problem:
\begin{equation}
\label{eq:loss_edge}
    \Delta{T}=\underset{\Delta{\hat{T}}}{\operatorname{argmin}} \sum_{i=1}^{N^{R}}||L_{i}^{\text{fit},\text{train}}-(L_{i}^{\text{fit},\text{test}}+\Delta{\hat{T}})||_2,
\end{equation}
where $L^{\text{fit, train}}$ and $L^{\text{fit, test}}$ denote the sets of landmarks used for fitting, selected from the face topologies defined in the train and test datasets, respectively. The term $N^{R}$ indicates the number of landmarks used in the fitting process. The refined prediction is calculated by adding $\Delta T$ to the initial prediction, and this refined result is used for evaluating the model’s performance.

%%%%%%%%%%%%%%%%%%%%%%%%%%%%%%%%%%%%%%%%%%%%%%%%%%%%%%%%
% Experiments
%%%%%%%%%%%%%%%%%%%%%%%%%%%%%%%%%%%%%%%%%%%%%%%%%%%%%%%%
\section{Experiments}
\label{sec:experiments}

%===================================================
\subsection{Implementation details}
\label{sec:implement_detail}

The input image height $H$ and width $W$ are both set to 192. The number of sampling points $N_t^P$ is set to $18 \times 18 = 324$ for $t = 0$ and 305 for $t = 1$. The number of vertices $N^V$ and landmarks $N^L$ are set to 1,220 and 68, respectively. To initialize the 3D dense landmarks $V_0$ and rotation $R_0$ of TRGv2, we selected a random sample from the ARKitFace training dataset~\cite{2023_tip_perspnet} and used its data. The initial correction parameters are set to $s_0 = 1$, $\tilde{\tau}_{0}^{x,\text{face}} = 0$, and $\tilde{\tau}_{0}^{y,\text{face}} = 0$. The loss weights $\lambda_{\text{head}}$, $\lambda_{\text{cam}}$, $\lambda_{\text{rot}}$, and $\lambda_{L}$ are set to 20, 2, 10, and 1.25, respectively. The performance of TRGv2 on the DD-Pose dataset~\cite{roth2019iv} was evaluated after training the model using only the DD-Pose training set. In contrast, all other TRGv2 results presented in this paper were evaluated after training exclusively on the ARKitFace training set~\cite{2023_tip_perspnet}. TRGv2 is trained end-to-end with a mini-batch size of 512 for a total of 30 and 200 epochs on the ARKitFace and DD-Pose datasets, respectively. We use the Adam optimizer~\cite{2015_Kingma} with an initial learning rate of $10^{-4}$, which is decreased by a factor of 10 after 20 epochs on the ARKitFace dataset and 180 epochs on the DD-Pose dataset. During training, data augmentation techniques such as random cropping, color jittering, image rotation, and horizontal flipping are applied to the training images. The training process takes approximately 9 and 30 hours on the ARKitFace and DD-Pose datasets, respectively, using two NVIDIA RTX 3090 GPUs.

%===================================================
\subsection{Datasets}
\label{sec:Datasets}

\textbf{ARKitFace}~\cite{2023_tip_perspnet} provides 6DoF head poses, dense 3D landmarks, and intrinsic camera parameters. It was collected in selfie scenarios, with camera-to-face distances ranging from 0.3 to 0.9 meters, resulting in images strongly affected by perspective distortion. Following the protocol of~\cite{2023_tip_perspnet}, we used 717,840 frames from 400 subjects for training and 184,884 frames from 100 subjects for testing.

\textbf{BIWI}~\cite{fanelli2013biwi} offers 6DoF head poses, a 3D neutral face mesh per subject, and camera intrinsics. Since BIWI does not provide ground-truth face meshes for each frame, our evaluation focuses on head pose only. BIWI serves exclusively as test data to assess the effectiveness of our method. We evaluated the performance of our proposed model by following the protocol used in~\cite{yang2019fsa, 2023_tip_perspnet}.

\textbf{DD-Pose}~\cite{roth2019iv} targets driver monitoring scenarios and consists of approximately 330,000 images captured under challenging conditions during driving, including occlusions, strong lighting, and truncations. The dataset provides 6DoF head poses and intrinsic parameters but lacks face geometry. To compensate for this, we manually annotated seven 2D landmarks corresponding to eyes, nose, mouth, and ears, and incorporated them into the training data. Following \cite{roth2023intrapose}, subjects 8, 19, and 23 were used for validation, subjects 3, 6, 10, 11, 14, 15, 16, and 17 for testing, and the remaining subjects for training. Performance on DD-Pose is evaluated via a dedicated server, which reports results separately for easy, moderate, and hard cases based on rotation and occlusion levels. For details on this categorization, please refer to \cite{roth2019iv}.

%===================================================
\subsection{Evaluation metrics}
\label{sec:eval_metrics}

For head rotation accuracy assessment, we follow the protocol used in previous studies~\cite{hsu2018quatnet, yang2019fsa, 2023_tip_perspnet, albiero2021img2pose} by reporting rotation errors separately for roll, pitch, and yaw. Additionally, we present the mean absolute error ($\text{MAE}_r$) and the geodesic error (GE)~\cite{cobo_pr24} for a more comprehensive evaluation of rotation performance. For DD-Pose, we also report the balanced mean angular error (BMAE)~\cite{schwarz2017driveahead, roth2023intrapose}. In driver monitoring scenarios, most frames are near-frontal, so models that perform well only on frontal poses can still achieve low overall $\text{MAE}_r$, which can hide poor performance on large head rotations. BMAE addresses this by weighting head rotation errors equally across the entire pose range, regardless of the sample distribution. To compute BMAE, all frames are first divided into fixed angular intervals based on their ground-truth head rotation. These intervals are defined using scalar rotation angles, which are obtained by converting the ground-truth head rotations into axis–angle representation. Then, $\text{MAE}_r$ values are computed within each interval, and their average is taken. The metric is defined as:
\begin{equation}
\label{eq:bmae}
    \text{BMAE}_{d,k}=\frac{d}{k}\sum_i \text{MAE}_r^{i,i+d}, \; i\in\{5j\,|\,j\in \mathbb{N}, 0 \le j \le \frac{k}{d} \},
\end{equation}
where $\text{MAE}_r^{i,i+d}$ denotes the $\text{MAE}_r$ measured within the interval $[i,i+d]$. Following \cite{roth2019iv, schwarz2017driveahead}, we set $d=5\degree$ and $k=75\degree$.

For evaluating the accuracy of head translation, we calculate errors along the $x$-, $y$-, and $z$-axes, represented as $t_x$, $t_y$, and $t_z$ errors, respectively. Similar to head rotation, we also report the mean absolute error for head translation, denoted as $\text{MAE}_t$. In this study, we align the head translation reference system on the BIWI dataset, as described in Section~\ref{sec:reference_align}, evaluate the model’s performance, and report the results accordingly. Following previous work~\cite{2023_tip_perspnet}, we utilize the average 3D distance ($\text{ADD}$) metric~\cite{hinterstoisser2013ADD} to provide a holistic evaluation that considers both rotation and translation:
\begin{equation}
\label{eq:ADD}
    \text{ADD}=\frac{1}{N^V}\sum_{n=1}^{N^V}\|(R_{2}^{mat}V_{n}^{*}+T_2) - (R^{*,mat}V_{n}^{*}+T^*)\|_{2}.
\end{equation}

To assess the 3D landmark prediction accuracy of our method, we also report the median and average distances between the estimated and ground-truth dense landmarks~\cite{2023_tip_perspnet}. The results are based on the final estimates ($V_2$, $R_2$, and $T_2$) from the final face regressor at $t=2$. Units for median, mean, translation error, and ADD are given in millimeters, while rotation errors are reported in degrees.

% table 1
%%%%%%%%%%%%%%%%%%%%%%%%
% table 1
% Ablation study for bidirectional interaction structure 
\begin{table}[t]
\scriptsize
\centering
\setlength\tabcolsep{1.0pt}
\def\arraystretch{1.1}
\caption{Ablation study of TRGv2 on the ARKitFace and BIWI datasets. All models were trained exclusively on the ARKitFace training data. The $\diamond$ symbol indicates values evaluated after reference system alignment.}
\label{tab:ablation for iterative inference}
\begin{tabular}{L{2.7cm}|C{0.9cm}C{0.9cm}C{0.9cm}C{0.9cm}|C{0.9cm}C{0.9cm}C{0.9cm}C{0.9cm}C{0.9cm}}
\specialrule{.1em}{.05em}{.05em}
\multirow{2}{*}{Method} & \multicolumn{4}{c|}{ARKitFace} & \multicolumn{5}{c}{BIWI}  \\ \cline{2-10}
{} & {Mean} & {$\text{GE}$} & {$\text{MAE}_t$ } & {$\text{ADD}$ } & {$\text{GE}$ } & {$\text{MAE}_t{\diamond}$} & {$\text{ADD}{\diamond}$} & \textcolor{gray}{$\text{MAE}_t$} & \textcolor{gray}{$\text{ADD}$} \\ \hline
{1-iter} & {1.63} & {1.98} & {3.71} & {8.96} & {6.28} & {12.31} & {32.16} & \textcolor{gray}{13.55} & \textcolor{gray}{31.43} \\
{2-iter (Ours)} & {1.59} & \textbf{1.76} & {3.64} & {8.76} & \textbf{5.73} & {11.33} & {29.25} & \textcolor{gray}{13.56} & \textcolor{gray}{31.35} \\
{3-iter} & {1.60} & \textbf{1.76} & {3.64} & {8.77} & {6.00} & \textbf{11.23} & \textbf{29.21} & \textcolor{gray}{13.42} & \textcolor{gray}{30.95} \\
\hline
{$T_t$-prediction} & {1.60} & {1.79} & \textbf{3.63} & \textbf{8.68} & {8.77} & {44.54} & {106.75} & \textcolor{gray}{41.64} & \textcolor{gray}{96.04} \\
% {Local-to-global~\cite{albiero2021img2pose}} & {1.60} & \textbf{1.76} & {3.66} & {8.81} & {5.74} & {12.16} & {31.55} & \textcolor{gray}{13.70} & \textcolor{gray}{30.82} \\

% {geometric baseline} & {1.57} & {1.79} & {20.13} & {42.05} & {5.38} & {49.20} & {125.29} & \textcolor{gray}{45.18} & \textcolor{gray}{112.41} \\
\hline
{Landmark-free baseline} & {-} & {2.04} & {4.05} & {9.79} & {6.38} & {13.11} & {33.32} & \textcolor{gray}{14.09} & \textcolor{gray}{32.63}  \\
{Grid-sampled baseline}  & \textbf{1.57} & {1.94} & {3.67} & {8.84} & {5.96} & {11.90} & {30.76} & \textcolor{gray}{\textbf{13.22}} & \textcolor{gray}{\textbf{30.43}} \\
\specialrule{.1em}{.05em}{.05em}
\end{tabular}
\end{table}
%%%%%%%%%%%%%%%%%%%%%%%%

%===================================================
\subsection{Ablation experiments}
\label{sec:ablation_experiments}

\textbf{Effectiveness of bidirectional interaction structure.} In this experiment, we examine the impact of explicit bidirectional interaction between the 6DoF head pose and face geometry. To this end, we analyze how the model’s performance varies with different numbers of interaction iterations between these two types of information. Specifically, we designed baseline models with one, two, and three interaction iterations and compared their performance. The one-iteration baseline simultaneously regresses the face geometry and head pose using $\mathcal{R}_1(\cdot)$ without any iterative refinement. In contrast, the two- and three-iteration models incorporate an iterative inference scheme that projects the predicted dense landmarks back onto the image features while keeping all other components consistent with the one-iteration baseline. 

As shown in Table~\ref{tab:ablation for iterative inference}, the evaluation results on the ARKitFace and BIWI datasets demonstrate that the two- and three-iteration models consistently outperform the one-iteration model across all metrics. This performance gain can be attributed to the reduction of ambiguity between the face geometry and the 6DoF head pose as the number of bidirectional interactions increases. 

According to the experimental results on BIWI dataset, although the two-iteration model slightly underperforms the three-iteration model in terms of translation accuracy, it significantly outperforms the three-iteration model in terms of rotation accuracy. Considering its balanced performance in both rotation and translation, we selected the two-iteration model as the final version of TRGv2.

\textbf{Use of correction parameters.} In this experiment, we investigate the rationale for estimating the correction parameters $c_t$ instead of directly estimating head translation $T_t$. To clarify this, we compare the performance of two models: the $T_t$-prediction baseline, which directly estimates head translation, and TRGv2. As shown in Table~\ref{tab:ablation for iterative inference}, while the $T_t$-prediction baseline achieves accurate head translation on the ARKitFace test data, its performance drops significantly on the BIWI dataset. This performance gap can be attributed to the different translation distributions in the ARKitFace and BIWI datasets.

% fig 5
%%%%%%%%%%%%%%%%%%%%%%%%
% fig 5
\begin{figure*}[t]
\centering
\includegraphics[width=1.0\linewidth]{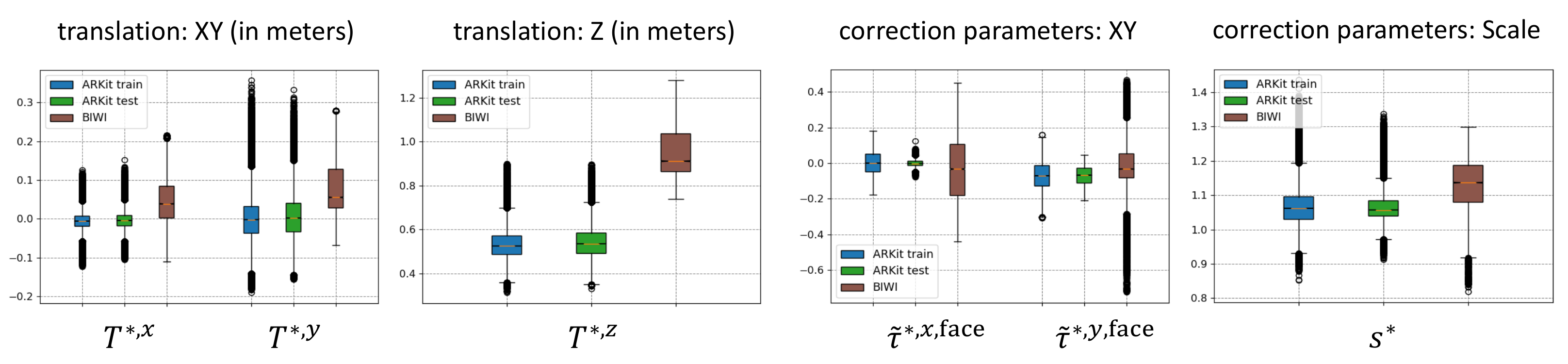}
\caption{Distributions of ground-truth translation and correction parameters in the ARKitFace and BIWI datasets. The colors blue, green, and brown represent the ARKitFace training data, ARKitFace test data, and the BIWI dataset, respectively. The symbol $*$ denotes the ground-truth.}
\label{fig:translation distribution}
\end{figure*}
%%%%%%%%%%%%%%%%%%%%%%%%

The first and second columns of Fig.~\ref{fig:translation distribution} illustrate the ground-truth head translation distributions for ARKitFace and BIWI. While the translation distribution in the ARKitFace training set closely aligns with its test set, it differs markedly from that of BIWI. This discrepancy is particularly noticeable in the $z$-axis translations, indicating substantial divergence between the ARKitFace training data and BIWI. This indicates that effective generalization from ARKitFace training data to BIWI requires the model to extrapolate the $z$-axis translation. However, as Table~\ref{tab:ablation for iterative inference} shows, this extrapolation is highly challenging for the direct translation estimation model.

The third and fourth columns of Fig.~\ref{fig:translation distribution} show the distribution of the ground-truth correction parameters for both datasets. Notably, the variation in the correction parameter distribution is much smaller than that of the translation distribution. These observations suggest that shifting the estimation target from $T_t$ to $c_t$ mitigates the distribution gap, thereby improving generalization to out-of-distribution data. This strategic redefinition enhances the model's generalizability, particularly for data that fall outside the training distribution, as evidenced in Table~\ref{tab:ablation for iterative inference}.

\textbf{The importance of utilizing facial geometry and the effectiveness of landmark-to-image projection technique.} For this experiment, we designed a landmark-free baseline that does not estimate facial geometry $\{V_t\}_{t=1}^2$. Due to the absence of facial geometry information, the landmark-free baseline cannot leverage landmark-to-image projection. Instead, it extracts grid-sampled features from $\phi$ and feeds them into the face regressor. However, given its significant structural differences from TRGv2, we mitigate this gap by additionally designing a grid-sampled baseline for incremental comparison. This grid-sampled baseline is identical to TRGv2 except that it does not employ the landmark-to-image projection, highlighting that the key difference from the landmark-free baseline is whether facial geometry is estimated.

% table 2
% table 2
  \begin{table}[t!]
    \scriptsize
    \centering
    \caption{Comparison of inference speed and GPU memory usage. ADD performance is measured on the BIWI dataset. The symbol $\diamond$ indicates values evaluated after reference system alignment. GPU usage is reported in GiB. TRG in this table denotes the conference version of TRG.}
    \label{tab:comparison_time_gpu}
    \begin{tabular}{L{2.7cm}|C{1.3cm}C{1.0cm}C{1.0cm}}
    \specialrule{.1em}{.05em}{.05em}
    Method & $\text{ADD}\ {\diamond}$ $\downarrow$ & FPS $\uparrow$ & GPU $\downarrow$ \\
    \hline
    PerspNet~\cite{2023_tip_perspnet} $\dagger$ & {31.88} & {115} & {1.17}  \\
    TRG~\cite{chun2024trg} & \textbf{29.11} & {162} & {0.21}  \\
    TRGv2 (Ours) & {29.25} & \textbf{208} & \textbf{0.18}  \\
    \specialrule{.1em}{.05em}{.05em}
    \end{tabular}
  \end{table}

% table 3
% table3
\begin{table}[t]
    \scriptsize
    \centering
    \caption{Comparison with previous methods for dense 3D landmark estimation on the ARKitFace test dataset.}
    \label{tab:comparison_geometry_arkit}
    \begin{tabular}{L{2.4cm}|C{1.0cm}C{1.0cm}}
        \specialrule{.1em}{.05em}{.05em}
        Method & $\text{Median}$ & $\text{Mean}$ \\ 
        \hline
        PRNet~\cite{feng2018prnet} & {1.97} & {2.05} \\
        3DDFA-v2~\cite{guo2020towards} & {2.35} & {2.31}  \\
        Deng \textit{et al.}~\cite{deng2019accurate} & {2.46} & {2.55}  \\
        JMLR~\cite{guo2022jmlr} $\dagger$ & {1.86} & {1.94} \\
        PerspNet~\cite{2023_tip_perspnet} & {1.72} & {1.76}  \\
        PerspNet~\cite{2023_tip_perspnet} $\dagger$ & {1.72} & {1.76}  \\
        MPC~\cite{xu2024multi} & {1.66} & {1.74} \\
        \hline
        TRGv2 (Ours) & \textbf{1.57} & \textbf{1.59}  \\
        \specialrule{.1em}{.05em}{.05em}
    \end{tabular}
  \end{table}

As shown in Table~\ref{tab:ablation for iterative inference}, the landmark-free baseline underperforms compared to the grid-sampled baseline. This result supports our hypothesis that landmark information should be integrated during the 6DoF head pose estimation process. Furthermore, our findings demonstrate that TRGv2 surpasses the grid-sampled baseline in head pose estimation, underscoring the effectiveness of our landmark-to-image projection strategy.

%===================================================
\subsection{Assessing model performance through reference system alignment}
\label{sec:reference_system_assess_model}

In this paragraph, we demonstrate that the reference system alignment method contributes to a more accurate assessment of model performance and plays an important role in model selection. To this end, we compare the cross-dataset evaluation results of the models presented in the ablation study, both before and after applying the reference system alignment method.

According to Table~\ref{tab:ablation for iterative inference}, without applying the reference system alignment, the grid-sampled baseline yields the lowest translation error. Moreover, the 3-iteration model shows significantly lower translation error and ADD compared to the 2-iteration model. However, after applying the reference system alignment, models that adopt the landmark-to-image strategy (i.e., the 2-iteration and 3-iteration models) actually outperform the grid-sampled baseline. Furthermore, the 2-iteration model performs nearly on par with the 3-iteration model in terms of translation estimation. We argue that the reference system alignment allows us to recognize that the 2-iteration model is indeed a strong performer. Without this alignment, we would not have been able to make this observation, which highlights the significance of the reference system alignment method.

% table 4
%%%%%%%%%%%%%%%%%%%%%%%%
% table 4
% SOTA, ARKitFace
\begin{table}[t]
\scriptsize
\centering
\setlength\tabcolsep{1.0pt}
\def\arraystretch{1.1}
\caption{Comparison with previous methods for 6DoF head pose estimation on the ARKitFace test dataset. The symbol $\dagger$ indicates a retrained model.}
\label{tab:detailed comparision on arkit}
\begin{tabular}{L{2.9cm}|C{0.7cm}C{0.7cm}C{0.7cm}C{0.9cm}C{0.9cm}C{0.7cm}C{0.7cm}C{1.0cm}C{1.0cm}C{1.0cm}}
\specialrule{.1em}{.05em}{.05em}
Method & Yaw & Pitch & Roll & \cellcolor{Gray}$\text{MAE}_r$ & \cellcolor{Gray}$\text{GE}$ & $t_x$ & $t_y$ & $t_z$ & \cellcolor{Gray}$\text{MAE}_t$ & \cellcolor{Gray}$\text{ADD}$ \\ \hline

img2pose~\cite{albiero2021img2pose, 2023_tip_perspnet} & {5.07} & {7.32} & {4.25} & \cellcolor{Gray}{5.55} & \cellcolor{Gray}{-} & {1.39} & {3.72} & {15.95} & \cellcolor{Gray}{7.02} & \cellcolor{Gray}{20.54} \\

Direct 6DoF Regress~\cite{2023_tip_perspnet} & {1.86} & {2.72} & {1.03} & \cellcolor{Gray}{1.87} & \cellcolor{Gray}{-} & {2.80} & {5.23} & {19.16} & \cellcolor{Gray}{9.06} & \cellcolor{Gray}{21.39} \\

Refined Pix2Pose~\cite{park2019pix2pose,2023_tip_perspnet} & {1.95} & {2.62} & {2.48} & \cellcolor{Gray}{2.35} & \cellcolor{Gray}{-} & {2.43} & {4.23} & {35.33} & \cellcolor{Gray}{14.00} & \cellcolor{Gray}{36.44} \\

JMLR~\cite{guo2022jmlr} $\dagger$ & {1.13} & {1.75} & {0.61} & \cellcolor{Gray}{1.16} & \cellcolor{Gray}{2.39} & {0.98} & {2.48} & {11.13} & \cellcolor{Gray}{4.86} & \cellcolor{Gray}{11.87}  \\

PerspNet~\cite{2023_tip_perspnet} & {0.98} & {1.43} & {0.55} & \cellcolor{Gray}{0.99} & \cellcolor{Gray}{-} & {1.00} & {2.41} & {9.73} & \cellcolor{Gray}{4.38} & \cellcolor{Gray}{10.30} \\

PerspNet~\cite{2023_tip_perspnet} $\dagger$ & {0.85} & {1.31} & {0.49} & \cellcolor{Gray}{0.88} & \cellcolor{Gray}{1.81} & {0.91} & {2.06} & {8.86} & \cellcolor{Gray}{3.95} & \cellcolor{Gray}{9.42} \\

MPC~\cite{xu2024multi} & {0.96} & {1.38} & {0.59} & \cellcolor{Gray}{0.98} & \cellcolor{Gray}{-} & {0.97} & {2.06} & {10.02} & \cellcolor{Gray}{4.35} & \cellcolor{Gray}{10.21} \\

\hline

{TRGv2} (Ours) & \textbf{0.85} & \textbf{1.26} & \textbf{0.52} & \cellcolor{Gray}\textbf{0.87} & \cellcolor{Gray}\textbf{1.76} & \textbf{0.80} & \textbf{1.90} & \textbf{8.23} & \cellcolor{Gray}\textbf{3.64} & \cellcolor{Gray}\textbf{8.76} \\

\specialrule{.1em}{.05em}{.05em}
\end{tabular}
\end{table}

%%%%%%%%%%%%%%%%%%%%%%%%

%===================================================
\subsection{Inference speed and GPU memory usage}
\label{sec:Inferece_gpu}

% table 5
%%%%%%%%%%%%%%%%%%%%%%%%
% table 5
% SOTA, BIWI
\begin{table}[!t]
\scriptsize
\centering
\setlength\tabcolsep{1.0pt}
\def\arraystretch{1.1}
\caption{Comparison with previous methods for 6DoF head pose estimation on the BIWI dataset. The models were evaluated on BIWI solely for testing purposes, without using it as training data. For reference system alignment, the head translation estimation was corrected as described in Section~\ref{sec:reference_align}.}
\label{table:comparison sota on biwi}
\begin{tabular}{L{3.0cm}|C{0.7cm}C{0.7cm}C{0.7cm}C{0.9cm}C{0.8cm}C{0.7cm}C{0.7cm}C{1.0cm}C{1.0cm}C{1.0cm}}
\specialrule{.1em}{.05em}{.05em}
Method & Yaw & Pitch & Roll & \cellcolor{Gray}$\text{MAE}_r$ & \cellcolor{Gray}$\text{GE}$ & $t_x$ & $t_y$ & $t_z$ & \cellcolor{Gray}$\text{MAE}_t$ & \cellcolor{Gray}$\text{ADD}$ \\ 

\hline
\hline

\multicolumn{3}{l}{w/o reference system alignment} \\
\hline

{Dlib~\cite{kazemi2014Dlib}} & {11.86} & {13.00} & {19.56} & \cellcolor{Gray}{14.81} & \cellcolor{Gray}{-} & {-} & {-} & {-} & \cellcolor{Gray}{-} & \cellcolor{Gray}{-} \\

{3DDFA~\cite{zhu2016_3ddfa}} & {5.50} & {41.90} & {13.22} & \cellcolor{Gray}{19.07} & \cellcolor{Gray}{-} & {-} & {-} & {-} & \cellcolor{Gray}{-} & \cellcolor{Gray}{-} \\

{EVA-GCN~\cite{xin2021eva}} & {4.01} & {4.78} & {2.98} & \cellcolor{Gray}{3.92} & \cellcolor{Gray}{-} & {-} & {-} & {-} & \cellcolor{Gray}{-} & \cellcolor{Gray}{-} \\

{HopeNet~\cite{ruiz2018hopenet}} & {4.81} & {6.61} & {3.27} & \cellcolor{Gray}{4.89} & \cellcolor{Gray}{9.53} & {-} & {-} & {-} & \cellcolor{Gray}{-} & \cellcolor{Gray}{-} \\

{QuatNet~\cite{hsu2018quatnet}} & {4.01} & {5.49} & {2.94} & \cellcolor{Gray}{4.15} & \cellcolor{Gray}{-} & {-} & {-} & {-} & \cellcolor{Gray}{-} & \cellcolor{Gray}{-} \\

{Liu \textit{et al.}~\cite{liu2019iccvw}} & {4.12} & {5.61} & {3.15} & \cellcolor{Gray}{4.29} & \cellcolor{Gray}{-} & {-} & {-} & {-} & \cellcolor{Gray}{-} & \cellcolor{Gray}{-} \\

{FSA-Net~\cite{yang2019fsa}} & {4.27} & {4.96} & {2.76} & \cellcolor{Gray}{4.00} & \cellcolor{Gray}{7.64} & {-} & {-} & {-} & \cellcolor{Gray}{-} & \cellcolor{Gray}{-} \\

{HPE~\cite{huang2020hpe}} & {4.57} & {5.18} & {3.12} & \cellcolor{Gray}{4.29} & \cellcolor{Gray}{-} & {-} & {-} & {-} & \cellcolor{Gray}{-} & \cellcolor{Gray}{-} \\

{WHENet-V~\cite{zhou2020whenet}} & {3.60} & {4.10} & {2.73} & \cellcolor{Gray}{3.48} & \cellcolor{Gray}{-} & {-} & {-} & {-} & \cellcolor{Gray}{-} & \cellcolor{Gray}{-} \\

{RetinaFace~\cite{deng2020retinaface}} & {4.07} & {6.42} & {2.97} & \cellcolor{Gray}{4.49} & \cellcolor{Gray}{-} & {-} & {-} & {-} & \cellcolor{Gray}{-} & \cellcolor{Gray}{-} \\

{FDN~\cite{zhang2020fdn}} & {4.52} & {4.70} & {2.56} & \cellcolor{Gray}{3.93} & \cellcolor{Gray}{-} & {-} & {-} & {-} & \cellcolor{Gray}{-} & \cellcolor{Gray}{-} \\

{MNN~\cite{valle2020mnn}} & {3.98} & {4.61} & {2.39} & \cellcolor{Gray}{3.66} & \cellcolor{Gray}{-} & {-} & {-} & {-} & \cellcolor{Gray}{-} & \cellcolor{Gray}{-} \\

{TriNet~\cite{cao2021trinet}} & \textbf{3.05} & {4.76} & {4.11} & \cellcolor{Gray}{3.97} & \cellcolor{Gray}{-} & {-} & {-} & {-} & \cellcolor{Gray}{-} & \cellcolor{Gray}{-} \\

{6DRepNet~\cite{hempel20226d}} & {3.24} & {4.48} & {2.68} & \cellcolor{Gray}{3.47} & \cellcolor{Gray}{-} & {-} & {-} & {-} & \cellcolor{Gray}{-} & \cellcolor{Gray}{-} \\

{Cao \textit{et al.}~\cite{caoECCV2022towards}} & {4.21} & {3.52} & {3.10} & \cellcolor{Gray}{3.61} & \cellcolor{Gray}{-} & {-} & {-} & {-} & \cellcolor{Gray}{-} & \cellcolor{Gray}{-} \\

{TokenHPE~\cite{zhang2023tokenhpe}} & {3.95} & {4.51}& {2.71} & \cellcolor{Gray}{3.72} & \cellcolor{Gray}{-} & {-} & {-} & {-} & \cellcolor{Gray}{-} & \cellcolor{Gray}{-} \\

{DAD-3DNet~\cite{martyniuk2022dad3dnet}} & {3.79} & {5.24}& {2.92} & \cellcolor{Gray}{3.98} & \cellcolor{Gray}{-} & {-} & {-} & {-} & \cellcolor{Gray}{-} & \cellcolor{Gray}{-} \\

{MFDNet~\cite{liu2021mfdnet}} & {3.40} & {4.68}& {2.77} & \cellcolor{Gray}{3.62} & \cellcolor{Gray}{-} & {-} & {-} & {-} & \cellcolor{Gray}{-} & \cellcolor{Gray}{-} \\

{Cobo \textit{et al.} (6D)~\cite{cobo_pr24}} & {4.58} & {4.65}& {2.71} & \cellcolor{Gray}{3.98} & \cellcolor{Gray}{7.30} & {-} & {-} & {-} & \cellcolor{Gray}{-} & \cellcolor{Gray}{-} \\

img2pose~\cite{albiero2021img2pose} & {4.57} & {3.55} & {3.24} & \cellcolor{Gray}{3.79} & \cellcolor{Gray}{7.10} & {-} & {-} & {-} & \cellcolor{Gray}{-} & \cellcolor{Gray}{-} \\

Direct 6DoF Regress~\cite{2023_tip_perspnet} & {16.49} & {14.03} & {5.81} & \cellcolor{Gray}{12.11} & \cellcolor{Gray}{-} & {62.36} & {85.01} & {366.52} & \cellcolor{Gray}{171.30} & \cellcolor{Gray}{562.38} \\

Refined Pix2Pose~\cite{park2019pix2pose,2023_tip_perspnet} & {5.75} & {5.06} & {11.23} & \cellcolor{Gray}{7.35} & \cellcolor{Gray}{-} & {16.82} & {21.30} & {255.36} & \cellcolor{Gray}{97.83} & \cellcolor{Gray}{356.32} \\

MICA~\cite{zielonka2022mica} & {5.40} & {7.17} & {3.80} & \cellcolor{Gray}{5.46} & \cellcolor{Gray}{-} & {9.32} & {13.66} & {60.13} & \cellcolor{Gray}{27.70} & \cellcolor{Gray}{68.03}  \\

JMLR~\cite{guo2022jmlr} $\dagger$ & {6.31} & {6.17} & {3.72} & \cellcolor{Gray}{5.40} & \cellcolor{Gray}{8.61} & {8.66} & {7.27} & {32.63} & \cellcolor{Gray}{16.19} & \cellcolor{Gray}{39.71}  \\

% PerspNet~\cite{2023_tip_perspnet} & {3.44} & {4.39} & {3.09} & {3.64} & \textbf{4.21} & \textbf{8.02} & {52.27} & {21.50} & {113.53} \\

PerspNet~\cite{2023_tip_perspnet} & {3.10} & {3.37} & {2.38} & \cellcolor{Gray}{2.95} & \cellcolor{Gray}{-} & \textbf{4.15} & \textbf{6.43} & {46.69} & \cellcolor{Gray}{19.09} & \cellcolor{Gray}{100.09} \\

PerspNet~\cite{2023_tip_perspnet} $\dagger$ & {3.06} & {3.48} & {2.12} & \cellcolor{Gray}{2.89} & \cellcolor{Gray}\textbf{5.61} & {7.82} & {8.08} & {28.39} & \cellcolor{Gray}{14.76} & \cellcolor{Gray}{33.68} \\

MPC~\cite{xu2024multi} & {3.67} & \textbf{3.26} & {2.16} & \cellcolor{Gray}{3.03} & \cellcolor{Gray}{-} & {-} & {-} & {-} & \cellcolor{Gray}{-} & \cellcolor{Gray}{-} \\

{TRGv2} (Ours) & {3.12} & {3.60} & \textbf{1.90} & \cellcolor{Gray}\textbf{2.87} & \cellcolor{Gray}{5.73} & {7.64} & {6.67} & \textbf{26.37} & \cellcolor{Gray}\textbf{13.56} & \cellcolor{Gray}\textbf{31.35} \\

\hline
\hline

\multicolumn{3}{l}{w/ reference system alignment}  \\
\hline

JMLR~\cite{guo2022jmlr} $\dagger$ & {6.31} & {6.17} & {3.72} & \cellcolor{Gray}{5.40} & \cellcolor{Gray}{8.61} & {3.77} & {4.59} & {31.65} & \cellcolor{Gray}{15.25} & \cellcolor{Gray}{35.54} \\

PerspNet~\cite{2023_tip_perspnet} $\dagger$ & \textbf{3.06} & \textbf{3.48} & {2.12} & \cellcolor{Gray}{2.89} & \cellcolor{Gray}\textbf{5.61} & {3.13} & \textbf{4.00} & {29.80} & \cellcolor{Gray}{12.31} & \cellcolor{Gray}{32.00} \\

{TRGv2} (Ours) & {3.12} & {3.60} & \textbf{1.90} & \cellcolor{Gray}\textbf{2.87} & \cellcolor{Gray}{5.73} & \textbf{2.99} & {4.13} & \textbf{26.89} & \cellcolor{Gray}\textbf{11.33} & \cellcolor{Gray}\textbf{29.25} \\

\specialrule{.1em}{.05em}{.05em}
\end{tabular}
\end{table}
%%%%%%%%%%%%%%%%%%%%%%%%

For a head pose estimator to be viable in production, it must satisfy not only predictive accuracy but also strict requirements on inference speed and GPU memory usage. To meet these constraints, we re-engineered the original conference version of TRG to create TRGv2, which delivers real-time performance even on low-end GPUs while maintaining high accuracy. Specifically, we reduced computational overhead by reducing the number of deconvolution layers from three to two and by reducing the number of TRG inference iterations from three to two. In addition, to minimize accuracy degradation, we relocated the sparse 2D landmark loss $\mathcal{L}_{L}$ so that it is applied to the features $\phi$. This re-engineering enables TRGv2 to achieve a $28.4 \%$ improvement in inference speed over TRG, as shown in Table~\ref{tab:comparison_time_gpu}. Compared with the previous state-of-the-art method PerspNet, TRGv2 achieves an $80.9 \%$ increase in speed while also reducing GPU memory usage by $84.6\%$. Moreover, TRGv2 outperforms PerspNet by a considerable margin in cross-dataset head pose estimation accuracy. These results demonstrate that TRGv2 combines superior accuracy, speed, and memory efficiency, making it significantly more suitable for real-world deployment than existing approaches.

% table 6
%%%%%%%%%%%%%%%%%%%%%%%%
% table 6
% DD-Pose 
\begin{table}[t]
\scriptsize
\centering
\setlength\tabcolsep{1.0pt}
\def\arraystretch{1.1}
\caption{Comparison of 6DoF head pose estimation methods on the DD-Pose validation and test sets.}
\label{table:comparison_sota_on_ddpose}
\begin{tabular}{L{1.8cm}|C{0.6cm}C{0.6cm}C{0.6cm}C{0.6cm}|C{0.6cm}C{0.6cm}C{0.6cm}C{0.6cm}|C{0.6cm}C{0.6cm}C{0.6cm}C{0.6cm}|C{0.45cm}C{0.45cm}C{0.45cm}C{0.45cm}}
\specialrule{.1em}{.05em}{.05em}
\multirow{2}{*}{Method} & \multicolumn{4}{c|}{BMAE $\downarrow$} & \multicolumn{4}{c|}{$\text{MAE}_r$ $\downarrow$} & \multicolumn{4}{c|}{$\text{MAE}_t$ $\downarrow$} & \multicolumn{4}{c}{recall $\uparrow$} \\ \cline{2-17}
{} & {all} & {e} & {m} & {h} & {all} & {e} & {m} & {h} & {all} & {e} & {m} & {h} & {all} & {e} & {m} & {h} \\

\hline
\hline
\multicolumn{3}{l}{validation set}  \\
\hline
{TRGv2 (Ours)} & {4.30} & {3.41} & {5.11} & {5.33} & {4.09} & {3.46} & {4.63} & {6.50} & {8.32} & {7.47} & {8.89} & {12.65} & {98} & {100} & {97} & {97} \\

\hline
\hline
\multicolumn{3}{l}{test set}  \\
\hline

{img2pose~\cite{albiero2021img2pose}} & {10.3} & {6.4} & {11.1} & {20.3} & {7.8} & {6.7} & {9.4} & {18.4} & {7.8K} & {7.7K} & {8.1K} & {8.4K} & {85} & {99} & {64} & {56} \\

{intrApose~\cite{roth2023intrapose}} & {5.8} & {4.2} & {6.2} & {9.5} & {4.8} & {3.9} & {5.9} & {8.9} & {25.0} & {21.5} & {29.4} & {41.5}& {97} & \textbf{100} & {93} & {93} \\

{TRGv2 (Ours)} & \textbf{4.8} & \textbf{3.4} & \textbf{5.3} & \textbf{8.8} & \textbf{4.2} & \textbf{3.6} & \textbf{4.9} & \textbf{7.0} & \textbf{18.3} & \textbf{16.1} & \textbf{21.8} & \textbf{22.1}& \textbf{99} & \textbf{100} & \textbf{98} & \textbf{97} \\

\hline

\specialrule{.1em}{.05em}{.05em}
\end{tabular}
\end{table}
%%%%%%%%%%%%%%%%%%%%%%%%

%===================================================
\subsection{Comparison with state-of-the-art methods}
\label{sec:comparison_sota}

In this experiment, we benchmarked our proposed method against existing approaches for 6DoF head pose estimation. The evaluation results on the ARKitFace, BIWI, and DD-Pose datasets are presented in Tables~\ref{tab:comparison_geometry_arkit}, \ref{tab:detailed comparision on arkit}, \ref{table:comparison sota on biwi}, and \ref{table:comparison_sota_on_ddpose}. Models that were retrained for this comparison are marked with the symbol $\dagger$.

\textbf{Evaluation on ARKitFace~\cite{2023_tip_perspnet}.} Img2pose directly estimates the 6DoF head pose from images without relying on explicit face geometry. In a similar direction, although MPC is fundamentally an optimization-based method, it also infers depth information directly from the image. However, the absence of explicit face geometry can increase face size ambiguity, which may degrade the accuracy of head pose estimation, as demonstrated in Table~\ref{tab:detailed comparision on arkit}.

\begin{figure}[t]
\centering
\includegraphics[width=1.0\linewidth]{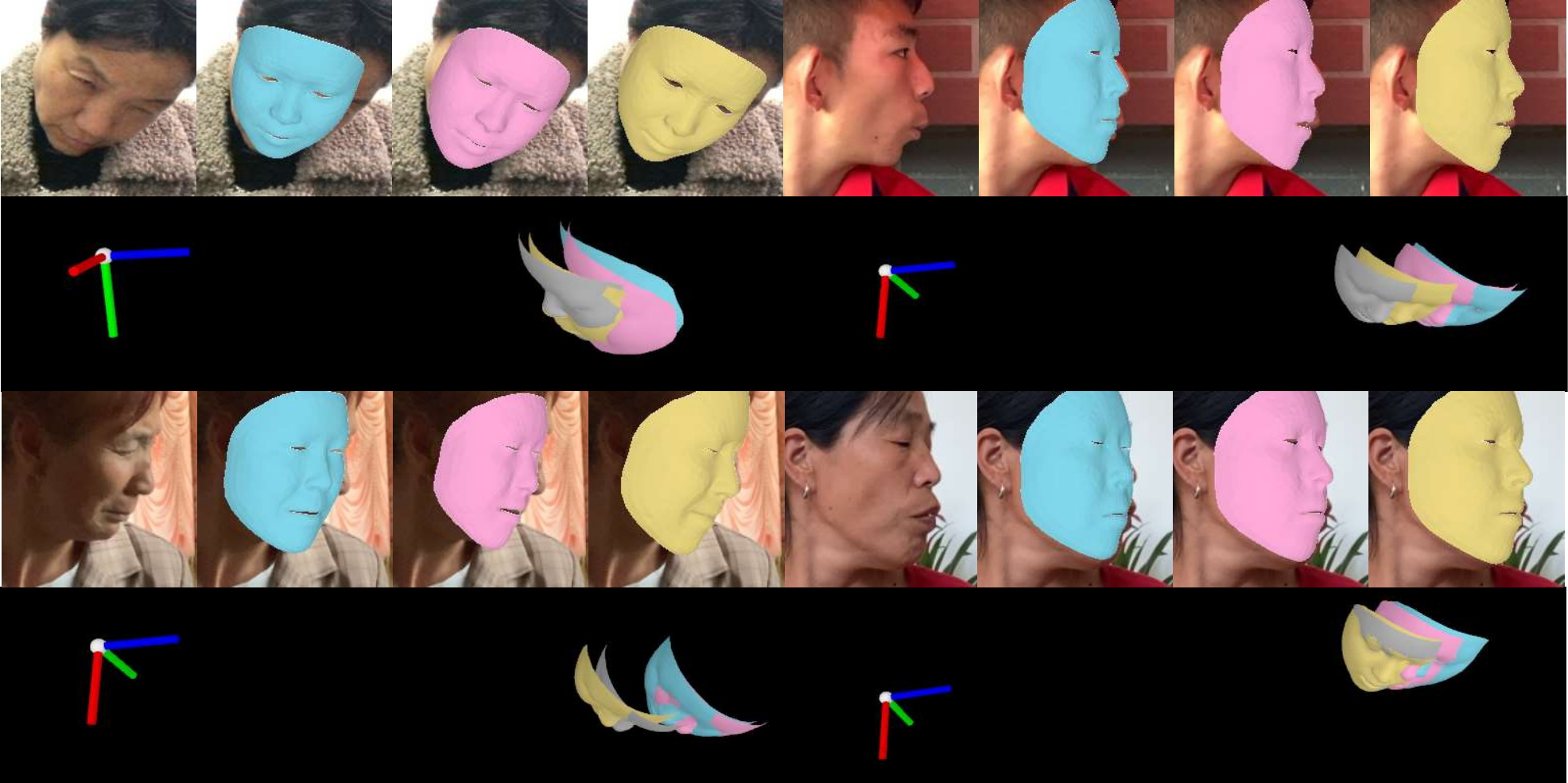}
\caption{Qualitative comparison on the ARKitFace dataset. The colors cyan, pink, gold, and gray represent JMLR, PerspNet, TRGv2, and the ground truth, respectively. The red, green, and blue axes respectively indicate the $X$, $Y$, and $Z$ axes of the camera coordinate system.}
\label{fig:qualitative_results_arkit}
\end{figure}

JMLR, PerspNet, and MPC do not incorporate head pose information during the face geometry inference process. As a result, the predicted face geometry, derived without considering head pose cues, is relatively inaccurate (Table~\ref{tab:comparison_geometry_arkit}). Consequently, methods that predict the 6DoF head pose based on such imprecise geometry yield lower accuracy (Table~\ref{tab:detailed comparision on arkit}). In contrast, TRGv2 explicitly integrates face geometry information into the head pose estimation process through its bidirectional interaction structure. TRGv2's bidirectional interaction, which involves a landmark-to-image projection strategy and an iterative inference based on this strategy, effectively enforces metric consistency between face geometry and 6DoF head pose. This consistency, in turn, contributes to improved head pose estimation performance. As shown in Table~\ref{tab:detailed comparision on arkit}, TRGv2 achieves state-of-the-art performance in head pose estimation. Furthermore, thanks to its depth-aware landmark prediction architecture, TRGv2 maintains stable facial landmark prediction accuracy in scenarios that are highly affected by perspective distortion, such as selfies (Table~\ref{tab:comparison_geometry_arkit}). Fig.~\ref{fig:qualitative_results_arkit} visually illustrates the performance of TRGv2 and existing models~\cite{guo2022jmlr, 2023_tip_perspnet} for head pose estimation and face landmark prediction. When the predicted geometries are aligned with the input images, they appear well-aligned. However, a stark contrast emerges when comparing the ground-truth geometry with the predicted geometries in 3D camera space: JMLR and PerspNet struggle to accurately predict the actual size of the human face, resulting in large translation errors.

% fig 6
%%%%%%%%%%%%%%%%%%%%%%%%
% fig 6
% the qualitative results: perspnet, jmlr, selfienet
%%%%%%%%%%%%%%%%%%%%%%%%

% fig 7
%%%%%%%%%%%%%%%%%%%%%%%%
% fig 7
% the qualitative results: perspnet, jmlr, selfienet
\begin{figure}[t]
\centering
\includegraphics[width=1.0\linewidth]{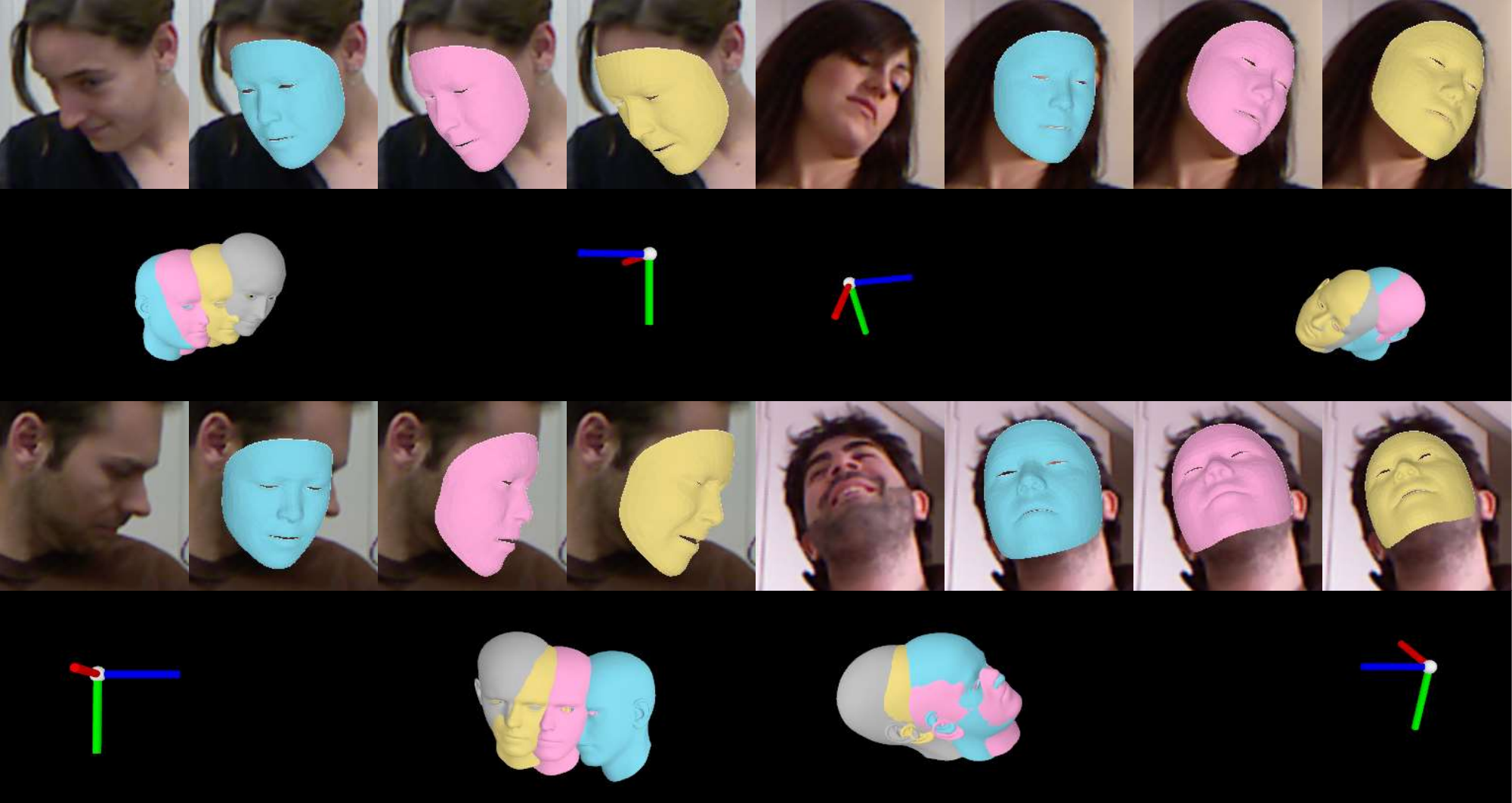}
\caption{Qualitative comparison on the BIWI dataset. The colors cyan, pink, gold, and gray represent JMLR, PerspNet, TRGv2, and the ground truth, respectively.}
\label{fig:qualitative_results_biwi}
\end{figure}
%%%%%%%%%%%%%%%%%%%%%%%%

\textbf{Evaluation on BIWI~\cite{fanelli2013biwi}.} We report both the unaligned performance following the conventional evaluation protocol and the performance measured after applying reference system alignment to enable a fair comparison. As shown in Table~\ref{table:comparison sota on biwi}, TRGv2 consistently outperforms existing optimization-based methods~\cite{zielonka2022mica, guo2022jmlr, 2023_tip_perspnet} in head translation estimation, both before and after reference system alignment. This superior performance is attributed to the design of TRGv2, which effectively leverages the synergy between face geometry and head translation. Fig.~\ref{fig:qualitative_results_biwi} qualitatively demonstrates TRGv2's strong head pose estimation capability. To illustrate how closely the predicted head pose matches the ground truth, we render the ground-truth neutral mesh using the predicted head pose.

\textbf{Evaluation on DD-Pose~\cite{roth2019iv}.} To evaluate our proposed model on the DD-Pose dataset, we employed bounding boxes extracted using RetinaFace~\cite{deng2020retinaface}. RetinaFace was fine-tuned on the DD-Pose training set, starting from publicly available pretrained weights, to enhance its face detection performance.

\begin{figure}[t]
\centering
\includegraphics[width=1.0\linewidth]{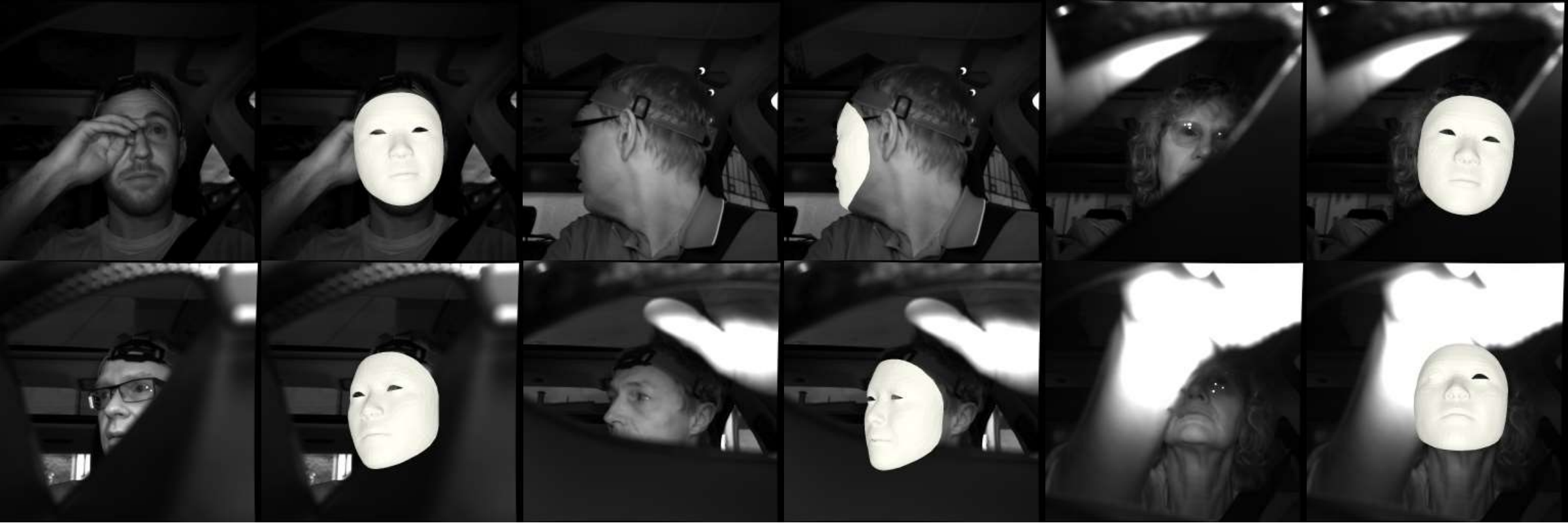}
\caption{Qualitative results of TRGv2 on the DD-Pose dataset.}
\label{fig:qualitative_results_ddpose}
\end{figure}

Prior to TRGv2, only img2pose and intrApose had been evaluated on the DD-Pose test set. Both methods estimate head pose directly from images without relying on explicit face geometry. TRGv2 outperforms img2pose and intrApose across all evaluation metrics (Table~\ref{table:comparison_sota_on_ddpose}). Notably, it achieves higher head pose estimation accuracy while also yielding better recall than these existing methods. In other words, TRGv2 maintains strong performance even when including more difficult cases such as severe occlusions and large head rotations. Fig.~\ref{fig:qualitative_results_ddpose} visually demonstrates that TRGv2 operates robustly even in challenging cases. These results suggest that TRGv2 is not only effective in controlled laboratory settings but also remains robust and accurate under challenging real-world conditions. To facilitate direct comparison with future methods, we also report TRGv2’s performance on the DD-Pose validation set in Table~\ref{table:comparison_sota_on_ddpose}.

% fig 8
%%%%%%%%%%%%%%%%%%%%%%%%
% fig 8
% the qualitative results: perspnet, jmlr, selfienet
%%%%%%%%%%%%%%%%%%%%%%%%

%===================================================
\subsection{Qualitative results on in-the-wild images}
\label{sec:comparison_sota}

\begin{figure}[t]
\centering
\includegraphics[width=0.8\linewidth]{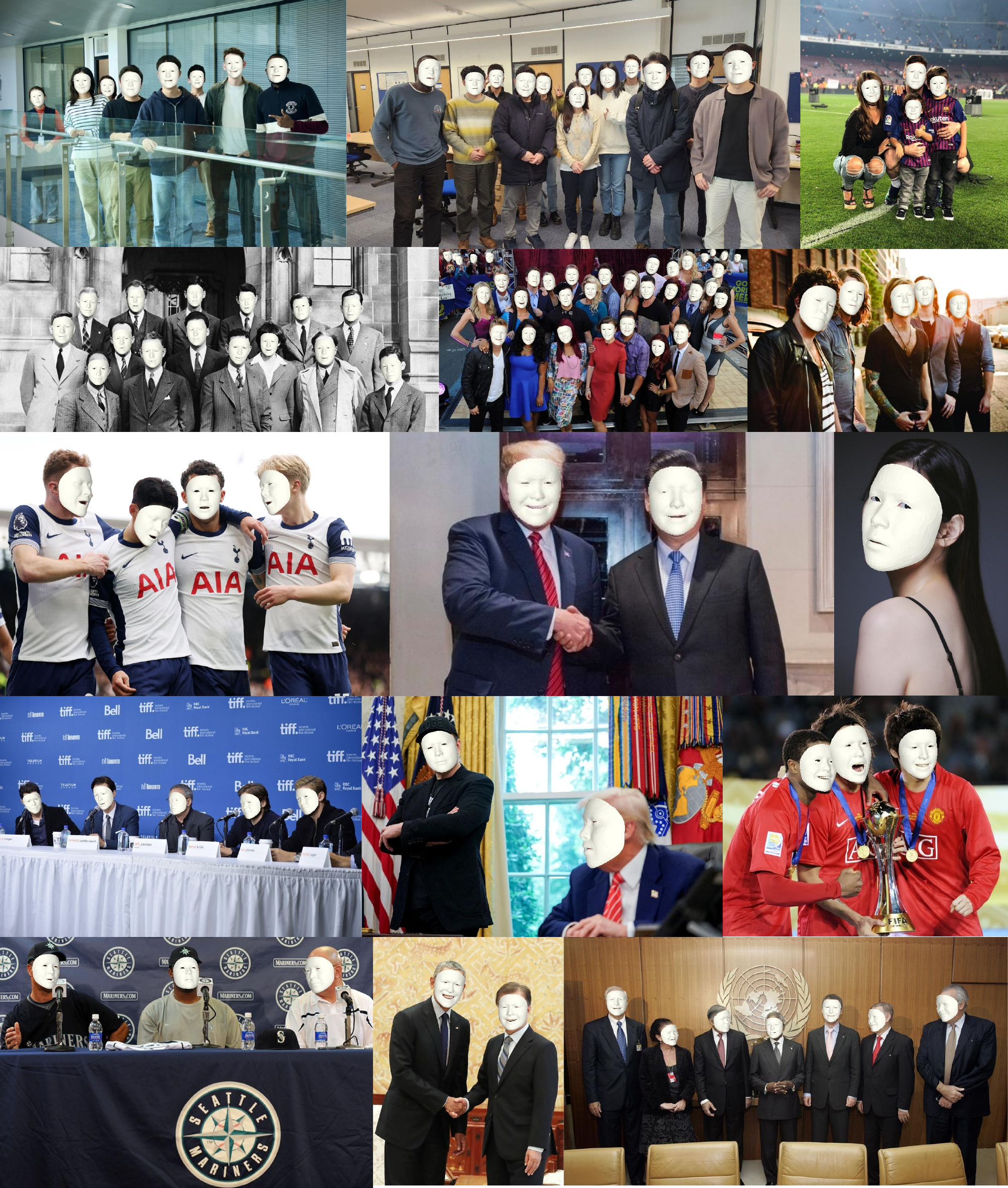}
\caption{Qualitative results of TRGv2 on in-the-wild images.}
\label{fig:itw_images}
\end{figure}

\begin{figure}[t]
\centering
\includegraphics[width=0.8\linewidth]{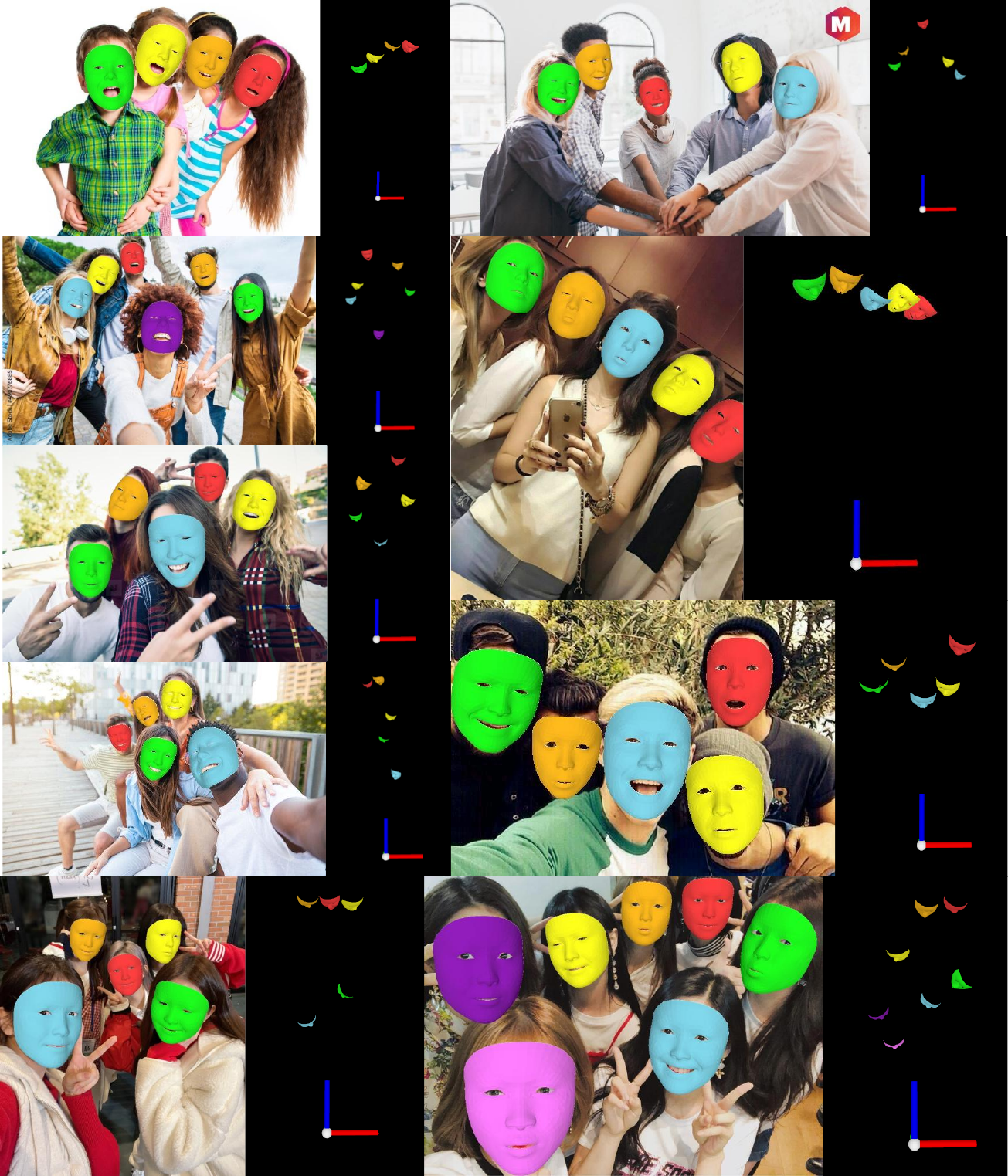}
\caption{Qualitative results of TRGv2 on in-the-wild images. For each example, the left side shows the face rendered on the input image, while the right side shows the face rendered in camera space. In camera space, the blue and red axes indicate the $Z$ and $X$ axes, respectively. Best viewed in color.}
\label{fig:itw_3d_space}
\end{figure}

\begin{figure}[t]
\centering
\includegraphics[width=1.0\linewidth]{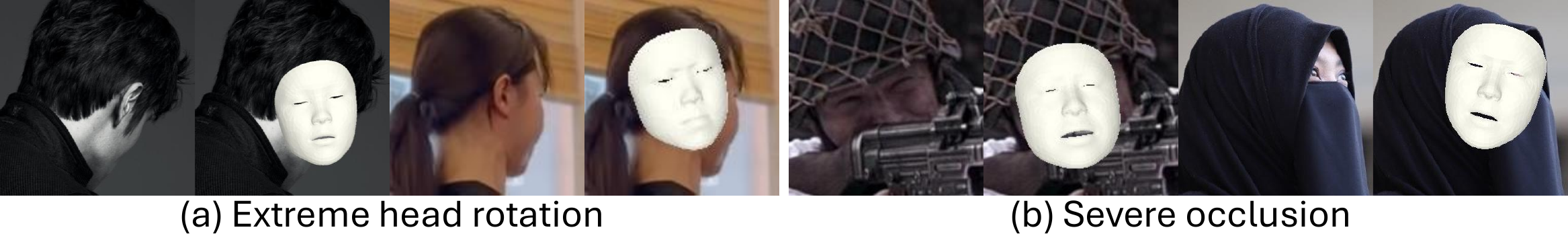}
\caption{Failure case. TRGv2 shows degraded performance when the test data significantly deviates from the training distribution.}
\label{fig:failure_case}
\end{figure}

To illustrate the effectiveness of our proposed method in real-world conditions, we provide additional qualitative results for images sourced from the internet, as shown in Figs.~\ref{fig:itw_images} and \ref{fig:itw_3d_space}. These results were inferred without knowledge of the camera intrinsics; instead, the focal length was simply approximated as the sum of the image’s width and height. Despite the lack of precise camera intrinsics, our method demonstrates reasonable performance in these unconstrained scenarios.
Nevertheless, as shown in Fig.~\ref{fig:failure_case}, our method does fail in some cases, such as those that include extreme head rotation or severe occlusion. The reason is that our method is a data-driven approach, and the dataset used to train it is mainly composed of frontal, unoccluded data.

% fig 10
%%%%%%%%%%%%%%%%%%%%%%%%
% fig 10
% In the wild, on image space
%%%%%%%%%%%%%%%%%%%%%%%%

% fig 11
%%%%%%%%%%%%%%%%%%%%%%%%
% fig 11
% the qualitative results: perspnet, jmlr, selfienet
%%%%%%%%%%%%%%%%%%%%%%%%

% fig 9
%%%%%%%%%%%%%%%%%%%%%%%%
% fig 9
% Failure case
%%%%%%%%%%%%%%%%%%%%%%%%

%===================================================
% \subsection{Limitations}
% \label{sec:limitation}

% As shown in Fig.~\ref{fig:failure_case}, TRGv2 exhibits performance degradation under extreme head rotations or severe facial occlusion.
% This is primarily because, although TRGv2 is a data-driven model, its training data is biased toward frontal and unoccluded faces, causing the network to overfit to this limited distribution. To ensure robust performance under full-range head rotations and severe occlusion, it is necessary to (i) construct a large-scale 6DoF head pose dataset that adequately covers such challenging scenarios, and (ii) develop training strategies that can effectively handle these difficult cases. We leave these directions as our future work.

%%%%%%%%%%%%%%%%%%%%%%%%%%%%%%%%%%%%%%%%%%%%%%%%%%%%%%%%
% Conclusion
%%%%%%%%%%%%%%%%%%%%%%%%%%%%%%%%%%%%%%%%%%%%%%%%%%%%%%%%
\section{Conclusion}
\label{sec:conclusion}

This study introduced TRGv2, a novel framework that predicts 6DoF head pose from a single image. Through extensive experiments, we demonstrated that the core design of TRGv2---explicit bidirectional interaction between the 6DoF head pose and dense 3D facial landmarks---is highly effective. We also showed that estimating correction parameters for the detected face bounding box significantly improves generalization performance in cross-dataset evaluations.

To the best of our knowledge, we are the first to highlight the need to reconsider existing cross-dataset evaluation protocols. Specifically, we identified a critical issue: differences in head center definitions across datasets can introduce bias in head translation evaluation. To address this hidden source of bias, we proposed a reference system alignment method for head translation. By applying this strategy, we were able to properly assess previously distorted model performance and make fair comparisons between TRGv2 and existing approaches.

Evaluations on the ARKitFace, BIWI, and DD-Pose datasets demonstrated that TRGv2 consistently outperforms existing state-of-the-art methods in head pose estimation. Specifically, TRGv2 reduced ADD by 0.66 mm ($7.01 \%$) on ARKitFace and 2.75 mm ($8.59 \%$) on BIWI, and achieved a 19.4 mm ($46.7 \%$) reduction in translation error on DD-Pose. These results suggest that TRGv2 is not only effective in controlled laboratory settings but also remains robust and accurate under challenging real-world conditions.

%However, our study has two limitations. First, TRG-v2 exhibits performance degradation in cases of extreme head rotations or severe facial occlusion. This is because TRG-v2 is a data-driven approach, and its training data is predominantly composed of frontal, unoccluded faces. Second, the proposed reference system alignment method is only applicable to datasets that provide 3D face geometry labels, as it calculates the difference in head center position based on face topology. To overcome these limitations, our future work will focus on constructing a large-scale 6DoF head pose dataset that adequately covers challenging scenarios and proposing a reference system alignment method that is free from face geometry.

Nevertheless, two limitations remain. First, TRGv2 tends to struggle with extreme head rotations or severe facial occlusion, as the training data predominantly consists of frontal, unoccluded faces. Second, the proposed reference system alignment method requires datasets that provide 3D face geometry labels, since it relies on face topology to calculate head center differences. Future work will therefore focus on constructing a large-scale 6DoF head pose dataset that adequately covers challenging scenarios, and on developing a geometry-independent alignment method.

%However, despite extensive experiments, we found that TRGv2 fails to make accurate predictions under full-range head rotations and in severe occlusion cases where facial features are barely visible. To overcome these limitations, it is necessary to build a large-scale 6-DoF head pose dataset that adequately covers such difficult scenarios and to develop models that can reliably operate under these conditions. We leave these directions as future work.

%%%%%%%%%%%%%%%%%%%%%%%%%%%%%%%%%%%%%%%%%%%%%%%%%%%%%%%%
% Etc
%%%%%%%%%%%%%%%%%%%%%%%%%%%%%%%%%%%%%%%%%%%%%%%%%%%%%%%%
\section*{Declaration of competing interest}

The authors declare that they have no known competing financial interests or personal relationships that could have appeared to influence the work reported in this paper.

\section*{Acknowledgement}

This work was partly supported by Institute of Information \& Communications Technology Planning \& Evaluation (IITP) grant funded by the Korea government (MSIT) (No. RS-2023-00219700, Development of FACS-compatible facial expression style transfer technology for digital human, 90\%), National Research Foundation of Korea (NRF) grant funded by the Korea government (MSIT) (No. NRF-2022R1F1A1066170, Physically valid 3D human motion reconstruction from multi-view videos, 10\%), and the research grant of Kwangwoon University in 2024.

%%%%%%%%%%%%%%%%%%%%%%%%%%%%%%%%%%%%%%%%%%%%%%%%%%%%%%%%%%%%
% References
%%%%%%%%%%%%%%%%%%%%%%%%%%%%%%%%%%%%%%%%%%%%%%%%%%%%%%%%%%%%
\bibliographystyle{elsarticle-num}   % pick the Elsevier numeric style
\bibliography{refs}

\end{document}